\documentclass[10pt,journal, compsoc]{IEEEtran}
\usepackage{amssymb}
\usepackage{amsthm}
\usepackage{amsmath}
\usepackage{mathrsfs}
\usepackage{stmaryrd}
\usepackage{graphicx,amsfonts,listings,xcolor,colortbl,subfigure}
\usepackage[linesnumbered,ruled,lined]{algorithm2e}
\usepackage{multirow}
\usepackage{tabu}
\usepackage{booktabs}
\usepackage{blkarray}
\usepackage{subfig}
\usepackage{arydshln}
\usepackage{setspace}
\usepackage{float}
\usepackage{epstopdf}
\usepackage[font=small,labelfont=bf]{caption}
\usepackage{verbatim}

\usepackage{array}
\usepackage{url}
\usepackage{verbatim}
\usepackage{cite}
\usepackage{diagbox}
\usepackage{bm}
\usepackage{cases}
\usepackage{ragged2e}

\renewcommand{\arraystretch}{1.3}

\newtheorem {theorem} {\bf Theorem}[]

\begin{document}

\title{Adaptive Collaborative Correlation Learning-based Semi-Supervised Multi-Label Feature Selection}

\author{Li~Yang,~Yanyong~Huang,~Dongjie~Wang,~Ke Li,~Xiuwen Yi,~Fengmao~Lv,~and~Tianrui~Li,\IEEEmembership{Senior Member,~IEEE}
\thanks{Li~Yang, Yanyong~Huang and Ke~Li are with the Joint Laboratory of Data Science and Business Intelligence, School of  Statistics,  Southwestern University of Finance and Economics, Chengdu 611130, China (e-mail: yangdali706@163.com; huangyy@swufe.edu.cn; likec@swufe.edu.cn);}
\thanks{Dongjie~Wang is with the Department of Electrical Engineering and Computer Science, University of Kansas, Lawrence, KS 66045, USA (e-mail: wangdongjie@ku.edu);}
\thanks{Xiuwen~Yi is with the JD Intelligent Cities Research and JD Intelligent Cities Business Unit, Beijing 100176, China (e-mail: xiuwenyi@foxmail.com);}
\thanks{Fengmao~Lv and Tianrui~Li are with the School of Computing and Artificial Intelligence, Southwest Jiaotong University, Chengdu 611756, China (e-mail: fengmaolv@126.com, trli@swjtu.edu.cn).}
}

\markboth{Journal of \LaTeX\ Class Files,~Vol.~14, No.~8, August~2021}%
{Shell \MakeLowercase{\textit{et al.}}: Adaptive Collaborative Correlation Learning-based Semi-Supervised Multi-Label Feature Selection}

\IEEEpubid{0000--0000/00\$00.00~\copyright~2021 IEEE}

\maketitle

\begin{abstract}
	Semi-supervised multi-label feature selection has recently been developed to solve the curse of dimensionality problem in high-dimensional multi-label data with certain samples missing labels. Although many efforts have been made, most existing methods use a predefined graph approach to capture the sample similarity or the label correlation. In this manner, the presence of noise and outliers within the original feature space can undermine the reliability of the resulting sample  similarity graph. It also fails to precisely depict the label correlation due to the existence  of unknown labels. Besides, these methods only consider the discriminative power of selected features, while neglecting their redundancy. In this paper, we propose an Adaptive Collaborative Correlation lEarning-based Semi-Supervised Multi-label Feature Selection (Access-MFS) method to address these issues. Specifically, a generalized regression model equipped with an extended uncorrelated constraint is introduced  to select discriminative yet irrelevant features and maintain consistency between predicted and ground-truth labels in labeled data, simultaneously. Then, the instance correlation and label correlation are integrated into the proposed regression model to adaptively learn both the sample similarity graph and the label similarity graph, which mutually enhance feature selection performance. Extensive experimental results demonstrate the superiority of the proposed Access-MFS over other state-of-the-art methods.		
\end{abstract}


\section{Introduction}
\IEEEPARstart{M}{ulti-label}Multi-label data, where each instance is associated with multiple labels, widely exists in real-world applications~\cite{ZhangandZhou2014,supervisedmulti-label1}. For example, a short video can be classified into several categorizes in the task of video classification~\cite{VideoCla},  a news article may be related to several subjects in text categorization~\cite{TextCla}, and a painting image is possible to be tagged into a number of genres in the image annotation task~\cite{ImageAnn}. In these applications, multi-label data is often represented in a high-dimensional feature space with redundant and noisy features, resulting in the ''curse of dimensionality'' problem and impacting the performance of subsequent tasks. Consequently, developing an effective method to reduce the dimensionality of multi-label data and thereby enhance downstream task performance has emerged as a pressing issue in practical applications.

\begin{figure*}[ht]
	\centering
	\includegraphics[width=1\textwidth]{Framework.pdf}
	\caption{The framework of the proposed adaptive collaborative correlation learning-based semi-supervised multi-label feature selection  method (Access-MFS).}
	\label{Framework}
\end{figure*}

Multi-label feature selection (MFS) deals with this issue by choosing a compact subset of discriminative features from the original high-dimensional feature space, thus eliminating redundant and noisy information~\cite{Qian2023,Hu2022}. In recent years, numerous  MFS methods have been developed, which can be broadly categorized into two groups. The first group converts  multi-label data into multiple independent single-label data and then applies the traditional feature selection method to the decomposed data~\cite{Li2017,Lapscore,Fu2021}. Typical feature selection methods for single-label data include the Laplacian Score~\cite{Lapscore}, spectral feature selection~\cite{Zhao2007} and group lasso-based feature selection~\cite{Wu2010}. Those methods usually assess the importance of features based on specific criteria while maintaining the local manifold structure of data, and subsequently select the top ranked features. However, those methods fail to consider the correlations among different labels, which could enhance the performance of feature selection. Instead of decomposing multi-label data into several single-label data, the second group of MFS methods directly construct a model from multi-label data  to facilitate feature selection~\cite{JianL2016,Braytee2017,JZhang2019,LFFS2022}. MIFS~\cite{JianL2016} is a typical method in the second group, which incorporates multi-label information into a low-dimensional subspace to identify discriminative features. Besides, Zhang et al. proposed using both local label correlations and global label correlations to guide the feature selection process~\cite{JZhang2023}. Moreover, Fan et al. have developed a collaborative learning framework for multi-label feature selection, which learns the label correlations and the feature correlations simultaneously~\cite{LFFS2022}. Despite the demonstrated efficacy of these proposed methods in feature selection, they share a common underlying assumption, i.e., all data has been comprehensively labeled. Unfortunately, obtaining large amounts of labeled data is expensive and time-consuming in real-world applications~\cite{Ma2012}. A more common scenario is that only a portion of the samples are labeled~\cite{semi-supervised1,Sun2021}. The aforementioned method based on a supervised way cannot be directly applied in the semi-supervised scenario. To address this issue, Chang et al. proposed a unified semi-supervised multi-label feature selection framework through the integration of sparse feature learning with label prediction~\cite{semi-supervised2}. Guo et al. developed a multi-label propagation mechanism to learn unknown labels, which was then integrated into an extended linear discriminant analysis for feature selection~\cite{LEDA2019}. Lv et al. combined adaptive graph structure learning and manifold learning to investigate the global structure of data and label correlations, and subsequently embedded them into the feature selection process~\cite{SFAM2021}. Constructing or learning the similarity graph is a promising technology for enhancing learning performance in MFS~\cite{WLiu2021}. However, existing semi-supervised MFS methods rely on predefined sample similarity graphs, while noise and outliers in the original feature space can undermine their reliability. Moreover, the label correlations encoded in a pre-computed manner may introduce inaccuracies and deficiencies, especially with large volumes of unlabeled data. Additionally, current methods that consider either label or sample correlations often fail to exploit their mutual reinforcement, thereby limiting the potential for enhanced feature selection.  Furthermore, these methods focus solely on the discriminative capability of selected features while ignoring their redundancy, ultimately degrading the performance of feature selection. 

To address these limitations, we present a novel MFS method for semi-supervised multi-label data, named as Adaptive Collaborative Correlation lEarning-based Semi-Supervised Multi-label Feature Selection (Access-MFS). Specifically, the proposed method Access-MFS first embeds the feature selection process into a generalized regression model equipped with an extended uncorrelated constraint, which enables the selection of discriminative yet irrelevant features while preserving consistency between predicted and ground-truth labels in labeled data. We then integrate the instance correlations and label correlations into the proposed regression model to adaptively learn the sample similarity graph and label similarity graph. The adaptive collaborative correlation learning module is capable of preserving the sample similarity from high-dimensional to low-dimensional space, guaranteeing that similar samples are assigned similar labels and that strong label correlations lead to consistent predicted labels, simultaneously. It yields more accurate and reliable sample and label similarity graphs than those obtained through the predefined graph method. Finally, an iterative optimization algorithm is developed to solve our model. Extensive experimental results demonstrate that the proposed Access-MFS is superior to  the state-of-the-art methods.

To sum up, the main contributions of this paper are as follows:  (\romannumeral1) To the best of our knowledge, we are the first to introduce adaptive collaborative correlation learning for feature selection on semi-supervised multi-label data. Our method can adaptively learn sample and label similarity graphs directly from the data, rather than relying on predefined graphs. The simultaneous learning of these two similarity graphs is integrated into the feature selection process, where they mutually enhance each other, leading to improved feature selection performance. (\romannumeral2) We propose a novel model for semi-supervised multi-label feature selection, incorporating a generalized regression module with an extended uncorrelated constraint and an adaptive collaborative correlation learning module. This model effectively selects features that are both discriminative and uncorrelated, while ensuring consistency between predicted labels and existing label data.  (\romannumeral3) We develop an efficient alternative optimization algorithm with guaranteed convergence to solve the proposed model. Comprehensive experiments are conducted on the real-world datasets to verify its superiority over state-of-the-art methods.

The remainder of this paper is organized as follows. Section 2 briefly reviews the related work about MFS.  We formulate the proposed method  Access-MFS in Section 3 and  provide an effective solution to this method in the following Section 4. Section 5 presents an analysis of the proposed algorithm's convergence behavior and time complexity. Section 6 conducts a series of experiments on eight real-world datasets. Finally, we conclude this article in Section 7.

\section{Related Work}\label{sec:Related work}
In this section, we briefly introduce some representative works about MFS. MFS can be classified into two categories based on whether it decomposes multi-labels. 

In the first category, multi-label data is transformed into multiple single-label datasets, upon which single-label feature selection methods are subsequently applied. Single-label feature selection methods are commonly classified into three groups: filter-based, wrapper-based, and embedded-based methods. The filter-based methods select informative features based on their importance, as defined by a specific evaluation criterion. Laplacian Score~\cite{Lapscore}, as a typical  filter-based approach, identifies representative features by utilizing the Laplacian Score metric, which quantifies the local preservative ability of the features. This type of method does not take into account the efficacy of the selected features on a downstream learning task, resulting in unsatisfactory performance. Wrapper-based methods typically employ a learning algorithm to assess the selected features until identifying the most suitable ones. Laporte et al. introduced a wrapper-based feature selection method that entails learning the ranking of features using support vector machines with a sparse constraint~\cite{Laporte2013}. The primary limitation of wrapper-based methods lies in their extensive time requirements and limited generalization capabilities. Embedded-based methods integrate feature selection into the model learning process, offering a middle ground between filter-based and wrapper-based methods. Liu et al. embedded the procedure of feature selection into the locally linear embedding algorithm based on the ${ \ell_{1}}$-norm reconstruction loss~\cite{YLiu2020}. These single-label feature selection methods ignore the label correlations, which will result in the inferior performance. 

Unlike the methods in the first category, the second category explores label correlations and constructs the feature selection model directly on multi-label data.  MIFS selects discriminative features by factorizing the label matrix to reveal label correlations and diminish the impact of imperfect label information~\cite{JianL2016}. In GLFS~\cite{JZhang2023}, the LASSO-based sparse learning is used to choose label-specific features that preserve groups, allowing for the simultaneous exploration of label-group and instance-group correlations. However, these methods assume that all samples are labeled, which may not hold true in real-world scenarios. More often, only a subset of samples are labeled. Multi-label feature selection within the semi-supervised learning paradigm is proposed to address this issue. For instance, Guo et al. devised a semi-supervised multi-label propagation method  to capture information from both labeled and unlabeled data simultaneously. Subsequently, the acquired label information is incorporated into an extended linear discriminant analysis module to select informative features~\cite{LEDA2019}. Xu et al. developed a semi-supervised multi-label feature selection algorithm that utilizes the probabilistic neighborhood similarities to learn the correlations between features and maintains consistent correlation information within the label space~\cite{SCFS2018}.  To explore the local manifold structures of both feature and label spaces in semi-supervised multi-label data, Xu et al.~\cite{SMLFS2021} uses a local logistic regression model  that incorporates feature graph regularization and label graph regularization. In addition, ${ \ell_{2,p}}$-norm is applied to the regression coefficient matrix to select important feature dimensions. Lv et al.~\cite{SFAM2021} introduced two modules to accurately represent the global structure of data and its label correlations: the adaptive global structure learning module, which maintains a global and sparse reconstruction of the structure, and the manifold learning module, focused on capturing the local structure and label correlations. These modules work synergistically to address their respective limitations, facilitating the selection of discriminative features. In SMDR-IC~\cite{SMDRIC2023}, Li et al. proposed a semi-supervised multi-label feature selection method that simultaneously utilizes label and sample correlations. By imposing binary hash constraints on the spectral embedding model, the method named SFS-BLL~\cite{SFS-BLL2023} has been proposed which identifies pseudo-labels for unlabeled data and utilizes a self-weighted sparse regression model to select discriminative features, leveraging both existing labels and learned pseudo-labels. Furthermore, Li et al. extended SMDR-IC and proposed SMDR-MRC~\cite{SMDR-MRC2024}, which decomposes the feature selection matrix into two matrices to respectively identify specific and common features, and introduces a non-zero correlation constraint to reduce redundant information between them. Using a shared subspace learning paradigm, Sheikhpour et al.~\cite{SMFS2025} decomposed the feature selection matrix into a label-specific weight matrix and a shared feature subspace term, while preserving both label relationships and feature correlations.Qing et al.~\cite{SFGR2025} proposed a semi-supervised multi-label feature selection method that selects label-relevant and non-redundant features using clustering, scoring, and ranking aggregation.

Although the aforementioned semi-supervised multi-label feature selection methodshave demonstrates certain effectiveness, several challenges still remain unaddressed. First, they use a predefined way to describe the sample similarity graph, which is easily influenced by noise and outliers. Second, these methods, which pre-compute label correlations based on partially known labels, fail to accurately capture the relationships between labels due to the large number of labels that remain unknown in a semi-supervised scenario. And existing methods fail to simultaneously consider the correlations between samples and between labels, overlooking how these two correlations can mutually enhance feature selection performance. Last, most approaches only take into account the discriminative capability of the chosen features and overlook their redundancy, which will limit the performance of feature selection. These issues are exhuastively discussed in our proposed Access-MFS method.

\section{Proposed method}\label{sec:proposed method}

\subsection{Notations}
In this paper, we denote matrices with boldface uppercase letters and vectors with boldface lowercase letters. For an arbitrary matrix ${\mathbf{M}} \in \mathbb{R}^{n \times p}$, $m_{ij}$ indicates the $i$th row and $j$th column entry of $\mathbf{M}$, $\mathbf{m}_{i.}$ and  $\mathbf{m}_{.j}$ denote its $i$th row and $j$th column, respectively. We use $\|\mathbf{M}\|_{F}=\sqrt{\sum_{i=1}^{n}\sum_{j=1}^{p}\left|m_{ij}\right|^{2}}$ to  denote the  Frobenius norm  of ${\mathbf{M}}$. The ${ \ell_{2,1}}$-norm of $\mathbf{M}$ is defined as $\Vert \mathbf{M} \Vert_{2,1}=\sum_{i=1}^n {\sqrt{\sum_{j=1}^{p} {m_{ij}^2}}}=\sum_{i=1}^n {\|\mathbf{m}_{i.}\|_{2}}$, where $\|\mathbf{m}_{i.}\|_{2}$ indicates the $\ell_{2}$-norm of the $i$th row vector $\mathbf{m}_{i.}$. Let $\operatorname{Tr}{(\mathbf{M})}$ and 
$\mathbf{M}^{T}$ respectively represent the trace and the transpose of ${\mathbf{M}}$. Besides, ${\mathbf{H}}={\mathbf{I}}-\frac{1}{n}\mathbf{1}_{n}\mathbf{1}_{n}^{T}$ is a centering matrix, where $\mathbf{I}$ is an identity matrix and ${\mathbf{1}_{n}}\in \mathbb{R}^{n \times 1}$ is a column vector of ones.

Let $\mathbf{X}=\left[\mathbf{X}_{l}, \mathbf{X}_u\right] \in \mathbb{R}^{d \times n}$ be a given data matrix with  $d$ dimensional features and $n$ instances, where $\mathbf{X}_{l} \in \mathbb{R}^{d \times n_1}$ is the labeled data, $\mathbf{X}_{u} \in \mathbb{R}^{d \times n_2}$ is the unlabeled data and ${n_1+n_2=n}$. The corresponding label matrix of $\mathbf{X}$ is denoted by  $\renewcommand{\arraystretch}{0.7}
\setlength{\arraycolsep}{0.1pt} 
\mathbf{Y} = \left[ \begin{array}{c} \mathbf{Y}_{l} \\ \mathbf{Y}_{u} \end{array} \right] \in \mathbb{R}^{n \times c}$ with $c$ categories, where $\mathbf{Y}_{l}\in \mathbb{R}^{n_1 \times c}$ and  $\mathbf{Y}_{u} \in \mathbb{R}^{n_2 \times c}$ denote the labels of $\mathbf{X}_{l}$ and $\mathbf{X}_u$, respectively. In $\mathbf{Y}_{l}$, ${y_{ij}=1}$ indicates  the $i$th instance is labeled as the $j$th class, otherwise it is 0. The label matrix $\mathbf{Y}_{u}$ of the unlabeled data is set to $0$. In semi-supervised multi-label feature selection, our aim is to select $k$ informative features from the given dataset, which  is much less than the total number of features in it. The framework of  Access-MFS is shown in Fig.~\ref{Framework}.

\subsection{Formulation of Access-MFS}
In order to learn a mapping function from samples with high dimensional features to the desired information (such as classification label or latent features), the least squares regression model has been extensively  applied in various machine learn tasks including classification, regression and dimension reduction~\cite{SXiang2012, XChen2017}. Given the input dataset $\mathcal{D}=\{(\mathbf{X},\mathbf{Y})\}$, where $\mathbf{X} \in \mathbb{R}^{d \times n}$ and 
$\mathbf{Y} \in \mathbb{R}^{n \times c}$ are the data matrix and the corresponding label matrix, respectively,  the traditional least squares regression model is defined as follows:
\begin{equation}\label{Nregression1}
	\begin{aligned}
		&\min _{\mathbf{W},\mathbf{b}}\|\mathbf{X}^{T} \mathbf{W}+\mathbf{1}_n \mathbf{b}^{T}-\mathbf{Y}\|_F^{2}+\lambda\Omega(\|\mathbf{W}\|),\\
	\end{aligned}
\end{equation}
where $\mathbf{W}\in \mathbb{R}^{d \times c}$ is a projection matrix, $\Omega(\|\mathbf{W}\|)$ indicates certain regularization on $\mathbf{W}$ and $\lambda$ is a regularization parameter. However, the abovementioned  model cannot be directly applied to semi-supervised multi-label feature selection.  Eq.~(\ref{Nregression1}) cannot be solved when only a subset of the labels in the label matrix  $\mathbf{Y}$ is known. Furthermore, to select the dicriminative features and avoid the trivial solution, a column full rank constraint  is typically  imposed on $\mathbf{W}$ in Eq.~(\ref{Nregression1}). This approach disregards  the redundancy of selected features. In order to deal with these two issues, we propose a generalized regression model with an extended uncorrelated constraint for semi-supervised multi-label feature selection, which is formulated  as follows:
\begin{equation}\label{New2}
	\begin{aligned}
		&\min_{\mathbf{W},\mathbf{b},\mathbf{F}}\|\mathbf{X}^{T} \mathbf{W}+\mathbf{1}_n \mathbf{b}^{T}-\mathbf{F}\|_F^{2}+{ \|{\mathbf{F}_{l}-\mathbf{Y}_{l}}\|_F^2}+\lambda\|\mathbf{W}\|_{2,1}\\
		&~{s.t.}~\mathbf{W}^{T}\mathbf{R} \mathbf{W}=\mathbf{I},
	\end{aligned}
\end{equation}
where $\renewcommand{\arraystretch}{0.7}
\setlength{\arraycolsep}{0.1pt} 
\mathbf{F} = \left[ \begin{array}{c} \mathbf{F}_{l} \\ \mathbf{F}_{u} \end{array} \right] \in \mathbb{R}^{n \times c}$ is a predicted label matrix and $\mathbf{W}^{T}\mathbf{R} \mathbf{W}=\mathbf{I}$ is an extended uncorrelated constraint. 
In Eq.~(\ref{New2}), $\mathbf{F}$ is comprised of the predicted label $\mathbf{F}_{l}$ for the labeled data $\mathbf{X}_{l}$ and $\mathbf{F}_{u}$  for the unlabeled data $\mathbf{X}_{u}$. We employ  the predicted label matrix $\mathbf{F}$ as an optimization variable in Eq.~(\ref{New2}),  simultaneously ensuring that the predicted label $\mathbf{F}_{l}$  is consistent with the ground-truth label $\mathbf{Y}_{l}$ of $\mathbf{X}_{l}$. Besides, $\ell_{2,1}$-norm is applied to $\mathbf{W}$ to induce row sparsity, facilitating feature selection. Then, we can select the top $k$ features by calculating the $\ell_2$ norm of each row in $\mathbf{W}$ and then sorting them in descending order.

To select the discriminative and uncorrelated features, we propose an extended uncorrelated constraint  on $\mathbf{W}$, denoted as $\mathbf{W}^{T}\mathbf{R} \mathbf{W}=\mathbf{I}$. Here,  $\mathbf{R}$ is defined as $\mathbf{X} \mathbf{H} \mathbf{X}^{T}+\lambda\mathbf{D}+\theta\mathbf{X}\mathbf{L}_{s}\mathbf{X}^T$, where
the matrix $\mathbf{D} \in \mathbb{R}^{d \times d}$ is diagonal, with the $i$th diagonal entry given by $\mathbf{D}(i, i)=\frac{1}{2 \sqrt{\|\mathbf{w}_{i.}\|_{2}^{2}+\epsilon}}$ ($\epsilon$ is a small constant to avoid division by 0). Additional, $\mathbf{L}_{s} =\mathbf{D}_{s}-\mathbf{S}$ represents the Laplacian matrix of instance similarity matrix $\mathbf{S}$, $\mathbf{D}_{s}$ is the degree matrix of $\mathbf{S}$, and $\theta$ is a trade-off parameter. The uncorrelated constraint we proposed consists of three terms obtained by expanding $\mathbf{R}$.  The first term aims to promote the orthonormality of the projected dimensions, which makes the low-dimensional instances to be uncorrelated. The second term is added to the first term to avoid the singularity of $\mathbf{X} \mathbf{H} \mathbf{X}^{T}$. The third term is advantageous in reducing the variance among samples within the same neighborhood under the graph structure. When $\lambda=\theta=0$, the proposed constraint reduces to the traditional uncorrelated constraint. The proposed  constraint is beneficial to select discriminative features that are also uncorrelated. Furthermore, it can simplify the optimization process of $\mathbf{W}$ and transforms $\mathbf{W}$ into a closed-form solution, as demonstrated in the optimization procedure.

The exploration of sample similarity structure and label correlations plays an important role in enhancing the performance of semi-supervised multi-label feature selection~\cite{LFFS2022,LEDA2019,SFAM2021,SMDRIC2023}. Many existing methods capture sample similarity and label correlations using predefined graph approaches. However, the presence of noise and outliers in the original feature space can compromise the accuracy of the resulting graph. Moreover, in semi-supervised settings with partially unknown labels, this approach cannot fully ascertain label correlations.  In this paper, we preserve the sample similarity structure and capture label correlations by adaptively learning both the instance similarity graph and the label similarity graph. This objective can be formally expressed as follows:
\begin{equation}
	\begin{aligned}\label{New3}
		&\min _{\mathbf{S},\mathbf{P}} \operatorname{Tr}(\mathbf{W}^{T}\mathbf{X}\mathbf{L}_{s}\mathbf{X}^{T}\mathbf{W})+\operatorname{Tr}(\mathbf{F}^{T} \mathbf{L}_{s} \mathbf{F})+\alpha \|\mathbf{S}\|_F^2\\
		\quad\quad&+\operatorname{Tr}(\mathbf{F} \mathbf{L}_{p} \mathbf{F}^{T})+\beta \|\mathbf{P}\|_F^2\\
		& {s.t.}~s_{i i}=0, s_{i j} \geq 0, \mathbf{S}\mathbf{1}_{n}=\mathbf{1}_{n}, p_{i i}=0, p_{i j} \geq 0, \mathbf{P}\mathbf{1}_{c}=\mathbf{1}_{c},
	\end{aligned}
\end{equation}
where $\mathbf{S}$ and $\mathbf{P}$ denote the instance and label similarity matrix, respectively, $\mathbf{L}_{s} =\mathbf{D}_{s}-\mathbf{S}$ and $\mathbf{L}_{p} =\mathbf{D}_{p}-\mathbf{P}$ are their corresponding Laplacian matrices, with $\mathbf{D}_{s}$ and $\mathbf{D}_{p}$ being diagonal matrices whose diagonal entries are  $\sum_{j=1}^{n} s_{i j}$ and $\sum_{j=1}^{c} p_{i j}$. Regularization parameters $\alpha$ and $\beta$ are used to control the sparsity of two similarity matrices. The first two terms of Eq.~(\ref{New3}) are designed to learn the instance similarity matrix from the perspectives of the feature space and the label space, respectively, while the forth term helps to maintain strong label correlations by encouraging related labels to have similar predicted values.  The Frobenius regularization terms ${\alpha \|\mathbf{S}\|_F^2}$ and ${\beta \|\mathbf{P}\|_F^2}$ prevent trivial solutions where each row of $\mathbf{S}$ or $\mathbf{P}$ collapses to a one-hot vector. By optimizing Eq.~(\ref{New3}), our framework collaboratively and adaptively learns both the instance similarity graph and the label similarity graph, rather than relying on predefined structures. This bidirectional coupling enables the two graphs to iteratively refine and reinforce each other, thus making them less vulnerable to noise, outliers, and label sparsity. As a result, the learned graphs provide more reliable structural priors, which substantially enhance the performance of semi-supervised multi-label feature selection.

By combining Eqs.~(\ref{New2}) and  (\ref{New3}) together, the proposed semi-supervised multi-label feature selection method Access-MFS is summarized as follows:
\begin{equation}\label{New4}
	\begin{aligned}
		&\min _{\Xi}\|\mathbf{X}^{T} \mathbf{W}+\mathbf{1}_n \mathbf{b}^{T}-\mathbf{F}\|_F^{2}+\lambda\|\mathbf{W}\|_{2,1}+{ \|{\mathbf{F}_{l}-\mathbf{Y}_{l}}\|_F^2}\\
		&\quad\quad+\theta (\operatorname{Tr}(\mathbf{W}^{T}\mathbf{X}\mathbf{L}_{s}\mathbf{X}^{T}\mathbf{W})
		+\operatorname{Tr}(\mathbf{F}^{T} \mathbf{L}_{s} \mathbf{F})+\alpha \|\mathbf{S}\|_F^2)\\
		&\quad\quad+\mu(\operatorname{Tr}(\mathbf{F} \mathbf{L}_{p} \mathbf{F}^{T})+\beta \|\mathbf{P}\|_F^2)\\
		&{s.t.}~\mathbf{W}^{T}\mathbf{R} \mathbf{W}=\mathbf{I}, s_{i i}=0, s_{i j} \geq 0, \mathbf{S}\mathbf{1}_{n}=\mathbf{1}_{n},p_{i i}=0, \\ 
		&~~~~~~~p_{i j} \geq 0,\mathbf{P}\mathbf{1}_{c}=\mathbf{1}_{c},
	\end{aligned}
\end{equation}
where $\Xi=\{\mathbf{W}, \mathbf{b},\mathbf{F}, \mathbf{S}, \mathbf{P}\}$, and $\theta$ and $\mu$ serve as trade-off hyper-parameters to control how strongly instance correlation and label correlation are captured during learning. These two regularization parameters $\alpha$ and $\beta$ can be automatically determined during the optimization process.

In our Access-MFS method, we integrate feature selection, the collaborative learning of instance similarity graphs and label similarity graphs, and the label learning into a unified framework, enabling these different learning tasks to promote each other. Access-MFS can offer two key benefits: First, the proposed extended regression model not only maintains consistency between the predicted labels and the ground-truth of the labeled data but also selects discriminative and uncorrelated features. Second, Access-MFS is capable of adaptively learning the instance and label similarity graphs that are more reliable than those generated by predefined methods. This collaborative learning of the two similarity-induced graphs can preserve the local structures of the feature space and label space concurrently.

\section{Optimization and Algorithm}\label{sec:optimization}
There are five variables that need to be optimized in Eq.~(\ref{New4}). Since these variables are interrelated, making it difficult to solve them simultaneously. To address this, an alternative iterative algorithm is proposed to solve the optimization problem by optimizing one variable at a time while keeping the others fixed.

Since there is no constraint on the variable $\mathbf{b}$, we can solve for it by setting the first-order derivate of  the objective function in Eq.~(\ref{New4}) w.r.t. $\mathbf{b}$ to 0, in accordance with the Karush-Kuhn-Tucker (KKT) conditions~\cite{Lagrange}. The optimal solution for $\mathbf{b}$ can be easily obtained as $\mathbf{b}=\frac{1}{n}(\mathbf{F}^{T}\mathbf{1}_{n}-\mathbf{W}^{T} \mathbf{X}\mathbf{1}_{n})$. By substituting $\mathbf{b}$ with the derived solution, Eq.~(\ref{New4}) can be reformulated as follows:
\begin{equation}\label{New5}
	\begin{aligned}
		&\min _{\mathbf{W},\mathbf{F},\mathbf{S},\mathbf{P}}\|\mathbf{H}\mathbf{X}^{T} \mathbf{W}-\mathbf{H}\mathbf{F}\|_{F}^{2}+\lambda\|\mathbf{W}\|_{2,1}+{ \|{\mathbf{F}_{l}-\mathbf{Y}_{l}}\|_F^2}\\
		&~~\quad\quad\quad+\theta (\operatorname{Tr}(\mathbf{F}^{T} \mathbf{L}_{s} \mathbf{F})+\operatorname{Tr}(\mathbf{W}^{T}\mathbf{X}\mathbf{L}_{s}\mathbf{X}^{T}\mathbf{W})
		\\
		&~~\quad\quad\quad+\alpha \|\mathbf{S}\|_F^2)+\!\mu(\operatorname{Tr}(\mathbf{F} \mathbf{L}_{p} \mathbf{F}^{T})+\beta \|\mathbf{P}\|_F^2)\\
		&{s.t.}~\mathbf{W}^{T}\mathbf{R} \mathbf{W}=\mathbf{I}, s_{i i}=0, s_{i j} \geq 0, \mathbf{S}\mathbf{1}_{n}=\mathbf{1}_{n},p_{i i}=0, \\
		\quad\quad&p_{i j} \geq 0,\mathbf{P}\mathbf{1}_{c}=\mathbf{1}_{c}.
	\end{aligned}
\end{equation}

In the following, we will describe the process for updating the variables $\mathbf{W}$, $\mathbf{F}$, $\mathbf{S}$ and $\mathbf{P}$ in Eq.~(\ref{New5}).

\subsection{Update $\mathbf{W}$ by Fixing Other Variables}
When other variables are fixed,  we can disregard the terms that do not involve $\mathbf{W}$ in Eq.~(\ref{New5}). Then, $\mathbf{W}$ can be obtained by solving the following problem:
\begin{equation}\label{New6}
	\begin{aligned}
		&\min _{\mathbf{W}}\|\mathbf{H}\mathbf{X}^{T} \mathbf{W}-\mathbf{H}\mathbf{F}\|_{F}^{2}\!+\!\lambda\|\mathbf{W}\|_{2,1}\!+\!\theta\operatorname{Tr}(\mathbf{W}^{T}\mathbf{X}\mathbf{L}_{s}\mathbf{X}^{T}\mathbf{W})\\
		&{s.t.}~\mathbf{W}^{T}\mathbf{R} \mathbf{W}=\mathbf{I}.
	\end{aligned}
\end{equation}

According to~\cite{FNie2010}, we have $\frac{\partial \|\mathbf{W}\|_{2,1}}{\partial \mathbf{W}}=2\mathbf{D}\mathbf{W}$. Then, Eq.~(\ref{New6}) can be transformed into  the following equivalent form:
\begin{equation}\label{New7}
	\begin{aligned}
		&\min _{\mathbf{W}}\|\mathbf{H}\mathbf{X}^{T} \mathbf{W}-\mathbf{H}\mathbf{F}\|_{F}^{2}+\lambda\operatorname{Tr}(\mathbf{W}^{T} \mathbf{D}\mathbf{W})\\
		&\quad\quad+\theta\operatorname{Tr}(\mathbf{W}^{T}\mathbf{X}\mathbf{L}_{s}\mathbf{X}^{T}\mathbf{W})\\
		&{s.t.}~\mathbf{W}^{T}\mathbf{R} \mathbf{W}=\mathbf{I}.
	\end{aligned}
\end{equation}

Based on the matrix calculus theory, Eq~(\ref{New7}) can be further derived as follows:
\begin{equation}\label{New8}
	\begin{aligned}
		&\min _{\mathbf{W}}\|\mathbf{H}\mathbf{X}^{T} \mathbf{W}-\mathbf{H}\mathbf{F}\|_{F}^{2}+\lambda\operatorname{Tr}(\mathbf{W}^{T} \mathbf{D}\mathbf{W})\\
		&\quad\quad+\theta\operatorname{Tr}(\mathbf{W}^{T}\mathbf{X}\mathbf{L}_{s}\mathbf{X}^{T}\mathbf{W})\\
		\Leftrightarrow&\min _{\mathbf{W}}\operatorname{Tr}(\mathbf{W}^{T}\mathbf{X}\mathbf{H}\mathbf{X}^{T}\mathbf{W}-2\mathbf{W}^{T}\mathbf{X}\mathbf{H}\mathbf{F}+\mathbf{F}^{T}\mathbf{H}\mathbf{F})\\&+\lambda\operatorname{Tr}(\mathbf{W}^{T}\mathbf{D}\mathbf{W})+\theta\operatorname{Tr}(\mathbf{W}^{T}\mathbf{X}\mathbf{L}_{s}\mathbf{X}^{T}\mathbf{W})\\
		\Leftrightarrow&\min _{\mathbf{W}}\operatorname{Tr}(\mathbf{W}^{T}\mathbf{R}\mathbf{W})
		-2\operatorname{Tr}(\mathbf{W}^{T}\mathbf{X}\mathbf{H}\mathbf{F})+\operatorname{Tr}(\mathbf{F}^{T}\mathbf{H}\mathbf{F})\\
	\end{aligned}
\end{equation}
Since the constraint $\mathbf{W}^{T}\mathbf{R} \mathbf{W}=\mathbf{I}$ and $\mathbf{F}$ is fixed, Eq.~(\ref{New8}) simplifies to the following problem.
\begin{equation}\label{New9}
	\max _{\mathbf{W}}\operatorname{Tr}(\mathbf{W}^{T}\mathbf{X}\mathbf{H}\mathbf{F})~{s.t.}~\mathbf{W}^{T}\mathbf{R} \mathbf{W}=\mathbf{I}.
\end{equation}

Let $\mathbf{A}=\mathbf{R}^{\frac{1}{2}}\mathbf{W}$ and $\mathbf{B}=\mathbf{R}^{-\frac{1}{2}}\mathbf{X}\mathbf{H}\mathbf{F}$. Then, Eq.~(\ref{New9}) is transformed into
\begin{equation}\label{New10}
	\max _{\mathbf{A}}\operatorname{Tr}(\mathbf{A}^{T}\mathbf{B})~{s.t.}~\mathbf{A}^{T}\mathbf{A}=\mathbf{I}.
\end{equation}

According to \cite{JinHuang2014}, the optimal solution for Eq.~(\ref{New10}) can be obtained as $\mathbf{A}=\mathbf{V}_{l}\mathbf{V}_{r}^{T}$, where $\mathbf{V}_{l}$ and $\mathbf{V}_{r}$ respectively denote the left and right singular matrices derived from the compact singular value decomposition (C-SVD)  of $\mathbf{B}$. Hence, the optimal solution for Eq.~(\ref{New9}) is derived as
\begin{equation}\label{New11}
	\mathbf{W}=\mathbf{R}^{-\frac{1}{2}}\mathbf{A}.
\end{equation}

In Eq. (\ref{New10}), as $\mathbf{R}$ contains $\mathbf{D}$, which is related to $\mathbf{W}$, we can update $\mathbf{W}$ and $\mathbf{D}$ alternatively to obtain the final optimal solution. The detailed procedure of solving problem~(\ref{New9}) is listed in Algorithm~\ref{NewAlgorithm1}.

\begin{algorithm}
	\caption{Algorithm to solve problem~(\ref{New9})}\label{NewAlgorithm1}
	\KwIn{Data matrix $\mathbf{X}$, centering matrix $\mathbf{H}$, predicted label matrix $\mathbf{F}$ and Laplacian matrix $\mathbf{L}_{s}$, parameters: $\lambda$ and $\theta$.
	}
	$\mathbf{Initialize:}$ $\mathbf{D}=\mathbf{I}_{d}$.
	
	\Begin{
		
		\While{not converged}{
			
			Compute $\mathbf{R}=\mathbf{X}\mathbf{H}\mathbf{X}^{T}+\lambda \mathbf{D}+\theta\mathbf{X}\mathbf{L}_{s}\mathbf{X}^T$;\\
			
			Compute $\mathbf{B}=\mathbf{R}^{-\frac{1}{2}}\mathbf{X}\mathbf{H}\mathbf{F}$;\\
			
			Compute $\mathbf{V}_{l}\sum \mathbf{V}_{r}^{T}=\mathbf{B}$ via C-SVD of $\mathbf{B}$;\\
			
			Compute $\mathbf{A}=\mathbf{V}_{l}\mathbf{V}_{r}^{T}$;\\
			
			Update $\mathbf{W}=\mathbf{R}^{-\frac{1}{2}}\mathbf{A}$;\\
			
			Update $\mathbf{D}=\operatorname{diag}(\frac{1}{2 \sqrt{\|\mathbf{w}_{1.}\|_{2}^{2}+\epsilon}}, \frac{1}{2 \sqrt{\|\mathbf{w}_{2.}\|_{2}^{2}+\epsilon}}, \cdots, \frac{1}{2 \sqrt{\|\mathbf{w}_{d.}\|_{2}^{2}+\epsilon}})$.\\
		}
	}
	\KwOut{The feature selection matrix $\mathbf{W}$.}
\end{algorithm}

\subsection{Update $\mathbf{F}$ by Fixing Other Variables}
By ignoring other fixed variables, we can solve  the following problem to obtain $\mathbf{F}$.
\begin{equation}\label{New12}
	\begin{aligned}
		&\min_{\mathbf{F}}\|\mathbf{H}\mathbf{X}^{T}\mathbf{W}-\mathbf{H}\mathbf{F}\|_{F}^{2}+\theta\operatorname{Tr}(\mathbf{F}^{T}\mathbf{L}_{s}\mathbf{F})+\mu\operatorname{Tr}(\mathbf{F} \mathbf{L}_{p} \mathbf{F}^{T})\\
		&\quad\quad+\|{\mathbf{F}_{l}-\mathbf{Y}_{l}}\|_F^2.
	\end{aligned}
\end{equation}

Let $\mathbf{U}$ be a diagonal matrix, where the $i$th diagonal entry $u_{ii}$ is defined as 1 if the $i$-th sample $\mathbf{x}_{.i}$ is labeled, and 0 otherwise. Then, we have $\|{\mathbf{F}_{l}-\mathbf{Y}_{l}}\|_F^2=\operatorname{Tr}((\mathbf{F}-\mathbf{Y})^{T}\mathbf{U}(\mathbf{F}-\mathbf{Y}))$. By following a similar deductive process as outlined in (\ref{New8}), problem~(\ref{New12}) can be transformed into the following equivalent trace form:
\begin{align}
	&\min_{\mathbf{F}}\operatorname{Tr}(\mathbf{F}^{T}\mathbf{Q}\mathbf{F}-2\mathbf{F}^{T}\mathbf{C})+\mu\operatorname{Tr}(\mathbf{F}\mathbf{L}_{p}\mathbf{F}^{T})\label{New13},
\end{align}
where $\mathbf{Q}=\theta\mathbf{L}_{s}+\mathbf{H}+\mathbf{U}$, $\mathbf{C}=\mathbf{H}\mathbf{X}^{T}\mathbf{W}+\mathbf{U}\mathbf{Y}$. 

Take the derivation of Eq.(\ref{New13}) w.r.t. ${\mathbf{F}}$ and set it to 0, we can obtain 
\begin{align}
	&\mathbf{Q}\mathbf{F}+\mu\mathbf{F}\mathbf{L}_{p}=\mathbf{C}\label{New14}
\end{align}
The optimal solution  ${\mathbf{F}}$ for Eq.~(\ref{New14})  can be obtained by using the proposed algorithm in~\cite{F-solve1}.

\subsection{Update $\mathbf{S}$ by Fixing Other Variables}
While $\mathbf{W}$, $\mathbf{F}$ and $\mathbf{P}$ are fixed, the optimization problem in Eq.~(\ref{New4}) is equivalent to solve:
\begin{equation}
	\begin{aligned}\label{New15}
		&\min _{\mathbf{S}} \operatorname{Tr}(\mathbf{W}^{T}\mathbf{X}\mathbf{L}_{s}\mathbf{X}^{T}\mathbf{W})+\operatorname{Tr}(\mathbf{F}^{T} \mathbf{L}_{s} \mathbf{F})+\alpha \|\mathbf{S}\|_F^2\\
		\Leftrightarrow &\min _{\mathbf{S}}\frac{1}{2}\sum_{i, j=1}^{n}\|\mathbf{W}^{T}\mathbf{x}_{.i}-\mathbf{W}^{T}\mathbf{x}_{.j}\|_{2}^{2} s_{i j}\\
		\quad\quad&+\frac{1}{2}\sum_{i, j=1}^{n}\|\mathbf{f}_{i.} -\mathbf{f}_{j.}\|_{2}^{2}{s_{ij}}+\alpha \|\mathbf{S}\|_F^2\\
		& {s.t.}~s_{i i}=0, s_{i j} \geq 0, \mathbf{S}\mathbf{1}_{n}=\mathbf{1}_{n}.
	\end{aligned}
\end{equation}

By setting a matrix $\mathbf{M}=(m_{ij})_{n \times n}$, where $m_{ij}=\frac{1}{2}\|\mathbf{W}^{T}\mathbf{x}_{.i}-\mathbf{W}^{T}\mathbf{x}_{.j}\|_{2}^{2}+\frac{1}{2}\|\mathbf{f}_{i.} -\mathbf{f}_{j.}\|_{2}^{2}$, Eq.~(\ref{New15}) is transformed into 
\begin{equation}\label{New16}
	\begin{aligned}
		&\min _{\mathbf{S}}\sum_{i, j=1}^{n} m_{i j}s_{i j}+\alpha\|\mathbf{S}\|_F^2\\
		&{s.t.}~s_{i i}=0, s_{i j} \geq 0, \mathbf{S}\mathbf{1}_{n}=\mathbf{1}_{n}.
	\end{aligned}
\end{equation}

Note that Eq.~(\ref{New16}) is independent for each row. Hence, for each $i$, we can rewritten Eq.~(\ref{New16}) by using the method of completing the square as follows:
\begin{equation}\label{New17}
	\begin{aligned}
		&\min _{\mathbf{s}_{i.}}\frac{1}{2}\|\mathbf{s}_{i.}+\frac{\mathbf{m}_{i.}}{2 \alpha}\|_{2}^{2} \\
		&{s.t.}~s_{i i}=0, s_{i j} \geq 0, \mathbf{s}_{i.}\mathbf{1}_n=1.
	\end{aligned}
\end{equation}

Then, the Lagrangian function of Eq.~(\ref{New17}) is written as:
\begin{equation}\label{New18}
	\begin{aligned}
		\mathcal{L}(\mathbf{s}_{i.}, \psi, \varphi)&=\frac{1}{2}\|\mathbf{s}_{i.}+\frac{\mathbf{m}_{i.}}{2 \beta}\|_{2}^{2}-\psi(\mathbf{s}_{i.}\mathbf{1}_n-1)-\bm{\varphi}^{T}\mathbf{s}_{i.}\\
		&=\frac{1}{2}\sum_{j=1}^{n}(s_{ij}\!+\!\frac{m_{ij}}{2 \alpha})^{2}\!-\!\psi(\sum_{j=1}^{n}s_{ij}-1)\!-\!\sum_{j=1}^{n}\varphi_{j}s_{ij},
	\end{aligned}
\end{equation}
where $\psi$ and $\bm{\varphi}=[\varphi_{1},\varphi_{2},\cdots,\varphi_{n}]^{T}$ are the Lagrangian multipliers. Based on the KKT conditions, we have ${s}_{i j}+\frac{m_{i j}}{2\alpha}-\psi-\varphi_{j}=0$ and $s_{i j}\varphi_{j}=0$. Then, we can get ${s}_{i j}=(\psi-\frac{m_{i j}}{2\alpha})_{+}$, where $(z)_{+}=max(z,0)$.

In practice, each sample tends to be similar to its neighbours. Therefore, we aim for $\mathbf{s}_{i.}$ to have $k_{s}$ nonzero values, indicating that the $i$th sample is similar to its $k_{s}$ closest neighbours. Assuming that $m_{i 1}, m_{i 2}, \cdots, m_{i n}$ are arranged in ascending order, we prefer ${s}_{i, k_{s}}>0$ and  ${s}_{i, k_{s}+1}=0$. As a result, we have $\alpha=\frac{m_{i, k_{s}+1}}{2\psi}$. Besides, according to the constraint $\mathbf{s}_{i.}\mathbf{1}_n=1$, we get $\psi=\frac{1}{k_{s}}+\frac{\sum_{j=1}^{k_{s}}m_{i j}}{2k_{s}\alpha}$. By substituting $\psi$ into the obtained $\alpha$, we can calculate   the regularization parameter as $\alpha=\frac{k_{s}m_{i,k_{s}+1}-\sum_{j=1}^{k_{s}}m_{ij}}{2}$. Then, we achieve the optimal solution ${s}_{i j}$ as follows:
\begin{equation}\label{New19}
	{s}_{i j}=\left\{\begin{array}{cl}
		\frac{m_{i, k_{s}+1}+m_{i j}}{k_{s} m_{i, k_{s}+1}-\sum_{h=1}^{k_{s}} m_{i h}} & j \leq k_{s}; \\
		0 & j>k_{s}.
	\end{array}\right.
\end{equation}

\subsection{Update $\mathbf{P}$ by Fixing Other Variables}
With other variables fixed, $\mathbf{P}$ can be optimized by solving the following problem:
\begin{equation}\label{New20}
	\begin{aligned}
		&\min_{\mathbf{P}}\operatorname{Tr}(\mathbf{F} \mathbf{L}_{p} \mathbf{F}^{T})+\beta \|\mathbf{P}\|_F^2\\
		\Leftrightarrow&\min_{\mathbf{P}}\frac{1}{2}\sum_{i, j=1}^{c}\|\mathbf{f}_{.i}-\mathbf{f}_{.j}\|_{2}^{2} p_{i j}+\beta \sum_{i=1}^{c}\|\mathbf{p}_{i.}\|_{2}^{2}\\
		&{s.t.}~p_{i i}=0, p_{i j} \geq 0,\mathbf{P}\mathbf{1}_{c}=\mathbf{1}_{c}.
	\end{aligned}
\end{equation} 

Let ${\mathbf{G}}$ be a ${c \times c}$ matrix with its $(i,j)$th entry  $g_{ij}=\frac{1}{2}\|\mathbf{f}_{.i}-\mathbf{f}_{.j}\|_{2}^{2}$. Then, Eq.~(\ref{New20}) is transformed into
\begin{equation}\label{New21}
	\begin{aligned}
		&\min_{\mathbf{P}}\sum_{i, j=1}^{c}g_{ij} p_{i j}+\beta \sum_{i=1}^{n}\|\mathbf{p}_{i.}\|_{2}^{2}\\
		&{s.t.}~p_{i i}=0, p_{i j} \geq 0,\mathbf{P}\mathbf{1}_{c}=\mathbf{1}_{c}.
	\end{aligned}
\end{equation} 

Given that the optimization of each row in $\mathbf{P}$ is independent, Eq.~(\ref{New21}) can be addressed by optimizing each individual subproblem as follows:
\begin{equation}\label{New22}
	\begin{aligned}
		&\min _{\mathbf{p}_{i.}}\frac{1}{2}\|\mathbf{p}_{i.}+\frac{\mathbf{g}_{i.}}{2 \beta}\|_{2}^{2} \\
		&{s.t.}~p_{i i}=0, p_{i j} \geq 0,\mathbf{p}_{i.}\mathbf{1}_c=1.
	\end{aligned}
\end{equation}

Following a similar procedure to that used in solving Eq.~(\ref{New17}), we can determine the regularization parameter $\beta=\frac{k_{p}g_{i,k_{p}+1}-\sum_{j=1}^{k_{p}}g_{ij}}{2}$, where $k_{p}$ denotes the number of nonzero elements in $\mathbf{p}_{i.}$. Furthermore, the optimal solution ${p}_{i j}$ is formulated as follows:
\begin{equation}\label{New23}
	p_{i j}=\left\{\begin{array}{cl}
		\frac{g_{i, k_{p}+1}+g_{i j}}{k_{p} g_{i, k+1}-\sum_{h=1}^{k_{p}} g_{i h}} & j \leq k_{p}; \\
		0 & j>k_{p}.
	\end{array}\right.
\end{equation}

Up to this point, we have obtained the updated expression of  $\mathbf{W}$, $\mathbf{F}$, $\mathbf{S}$ and $\mathbf{P}$. The detailed optimization procedure of Access-MFS is summarized in Algorithm~\ref{NewAlgorithm2}.

\begin{algorithm}
	\caption{Access-MFS}\label{NewAlgorithm2}
	\setlength{\arraycolsep}{0.1pt}
	\renewcommand{\arraystretch}{0.8}
	\KwIn{ \begin{enumerate}
			\item The data matrix $\mathbf{X}=\left[\mathbf{X}_{l}, \mathbf{X}_u\right]\in \mathbb{R}^{d \times n}$, and the centering matrix $\mathbf{H}$;
			\item The label matrix $\mathbf{Y}=\left[ \begin{array}{c} \mathbf{Y}_{l} \\ \mathbf{0} \end{array} \right]\in \mathbb{R}^{n \times c};$
			\item The trade-off parameters $\lambda,\theta$ and $\mu$.
		\end{enumerate}
	}
	$\bf{Initialization:}$ $\mathbf{F}=\mathbf{Y}$, randomly initialize $\mathbf{W}$, $\mathbf{S}$ and $\mathbf{P}$ are initialized by  Eq.~(\ref{New19}) and Eq.~(\ref{New23}), respectively.
	
	\Begin
	{
		\While{not converged}{
			Update $\mathbf{W}$ by using Algorithm~\ref{NewAlgorithm1};\\
			Update $\mathbf{F}$ by solving Eq.~(\ref{New14});\\
			Update $\mathbf{S}$ via Eq.~(\ref{New19});\\
			Update $\mathbf{P}$ via Eq.~(\ref{New23});\\
		}
	}
	\KwOut{Sort the scores of all features computed by the $\ell_{2}$-norm of the rows of $\mathbf{W}$ in descending order and select the top $k$ ranked features.}
\end{algorithm}

\section{Convergence and Complexity Analysis}\label{sec:Convergence and Complexity Analysis}
In this section, we provide a theoretical analysis of the proposed algorithm Access-MFS, which  includes both the convergence analysis and the complexity analysis.

\subsection{Convergence Analysis}\label{sec:Convergence}
Due to the objective function in Eq.~(\ref{New5}) not being convex w.r.t. four variables $\mathbf{W}$, $\mathbf{F}$, $\mathbf{S}$ and $\mathbf{P}$ simultaneously, we address this by dividing it into four sub-objective functions, namely, Eqs.~(\ref{New6}), (\ref{New12}), (\ref{New15}), and (\ref{New20}), each of which is separately solved  with other variables fixed. Therefore, the convergence of Algorithm~\ref{NewAlgorithm2} can be demonstrated by proving that each sub-objective function decreases monotonically. We first present the following theorem to prove the monotonic  decrease of the objective function in Eq.~(\ref{New6}) when updating $\mathbf{W}$ with $\mathbf{F}$, $\mathbf{S}$ and $\mathbf{P}$ are fixed.

\begin{theorem}\label{NewTheorem1}
	The objective function in Eq.~(\ref{New6}) monotonically decreases until convergence when  $\mathbf{W}$ is updated using the rules outlined in Algorithm~\ref{NewAlgorithm2}, with the other variables fixed.  
\end{theorem}

\begin{proof}
	Let $\varUpsilon(\mathbf{W}^{(t)},\mathbf{D}^{(t)})=\|\mathbf{H}\mathbf{X}^{T} \mathbf{W}^{(t)}-\mathbf{H}\mathbf{F}\|_{F}^{2}+\lambda\operatorname{Tr}({\mathbf{W}^{(t)}}^{T} \mathbf{D}^{(t)}\mathbf{W}^{(t)})+\theta\operatorname{Tr}({\mathbf{W}^{(t)}}^{T}\mathbf{X}\mathbf{L}_{s}\mathbf{X}^{T}\mathbf{W}^{(t)})$ represent the objective  value in the $t$th iteration of Eq.~(\ref{New7}). According to \cite{JinHuang2014}, during the alternate updating of $\mathbf{W}$ and $\mathbf{D}$ using Algorithm~\ref{NewAlgorithm1}, the following inequalities must hold for the ($t$+1)th iteration:
	\begin{equation}\label{New24}
		\begin{aligned}
			\varUpsilon(\mathbf{W}^{(t+1)},\mathbf{D}^{(t)}) \leq \varUpsilon(\mathbf{W}^{(t)},\mathbf{D}^{(t)})
		\end{aligned}
	\end{equation}
	
	Besides, by means of  the property $\|\mathbf{u}\|_{2}-\frac{\|\mathbf{u}\|_{2}^{2}}{2\|\mathbf{v}\|_{2}} \leq\|\mathbf{v}\|_{2}-\frac{\|\mathbf{v}\|_{2}^{2}}{2\|\mathbf{v}\|_{2}}$ for any two nonzero vectors $\mathbf{u}$ and $\mathbf{v}$  as proposed in~\cite{FNie2010}, we have 
	\begin{align}
		&\|\mathbf{w}^{(t+1)}_{i.}\|_{2}-\frac{\|\mathbf{w}^{(t+1)}_{i.}\|_{2}^{2}}{2\|\mathbf{w}^{(t)}_{i.}\|_{2}} \leq\|\mathbf{w}^{(t)}_{i.}\|_{2}-\frac{\|\mathbf{w}^{(t)}_{i.}\|_{2}^{2}}{2\|\mathbf{w}^{(t)}_{i.}\|_{2}}\nonumber\\
		\Rightarrow&\lambda\sum_{i=1}^{d}\|\mathbf{w}^{(t+1)}_{i.}\|_{2}-\lambda\sum_{i=1}^{d}\frac{\|\mathbf{w}^{(t+1)}_{i.}\|_{2}^{2}}{2\|\mathbf{w}^{(t)}_{i.}\|_{2}} \leq\nonumber\\&\lambda\sum_{i=1}^{d}\|\mathbf{w}^{(t)}_{i.}\|_{2}-\lambda\sum_{i=1}^{d}\frac{\|\mathbf{w}^{(t)}_{i.}\|_{2}^{2}}{2\|\mathbf{w}^{(t)}_{i.}\|_{2}}\nonumber\\
		\Rightarrow&\lambda\|\mathbf{W}^{(t+1)}\|_{2,1}-\lambda\operatorname{Tr}({\mathbf{W}^{(t+1)}}^{T}\mathbf{D}^{(t)}\mathbf{W}^{(t+1)}) \leq\nonumber\\& \lambda\|\mathbf{W}^{(t)}\|_{2,1}-\lambda\operatorname{Tr}({\mathbf{W}^{(t)}}^{T}\mathbf{D}^{(t)}\mathbf{W}^{(t)})\label{New25}.
	\end{align}
	By combing the inequalities~(\ref{New24}) and (\ref{New25}), it is easy to obtain
	\begin{equation}\label{26}
		\begin{aligned}
			&\|\mathbf{H}\mathbf{X}^{T} \mathbf{W}^{(t+1)}-\mathbf{H}\mathbf{F}\|_{F}^{2}+\lambda\|\mathbf{W}^{(t+1)}\|_{2,1} \\&+\theta\operatorname{Tr}({\mathbf{W}^{(t+1)}}^{T}\mathbf{X}\mathbf{L}_{s}\mathbf{X}^{T}\mathbf{W}^{(t+1)})\\
			&\leq \|\mathbf{H}\mathbf{X}^{T} \mathbf{W}^{(t)}-\mathbf{H}\mathbf{F}\|_{F}^{2}+\lambda\|\mathbf{W}^{(t)}\|_{2,1} \\&+\theta\operatorname{Tr}({\mathbf{W}^{(t)}}^{T}\mathbf{X}\mathbf{L}_{s}\mathbf{X}^{T}\mathbf{W}^{(t)}).
		\end{aligned}
	\end{equation}
	
	Hence, the objective function in Eq.~(\ref{New5}) monotonically decreases by Algorithm~\ref{NewAlgorithm2} in each iteration.
\end{proof}

As Eq.~(\ref{New12}) is equivalent to Eq.~(\ref{New13}), we only need to demonstrate that the objective function in Eq.(\ref{New13}) with respect to ${\mathbf{F}}$ is a monotonically decreasing function. It is easy to verify that the matrix $\mathbf{Q}$ and the Laplacian matrix $\mathbf{L}_{p}$ are both positive semidefinite. Consequently, the quadratic forms $\mathbf{F}^{T}\mathbf{Q}\mathbf{F}-2\mathbf{F}^{T}\mathbf{C}$ and $\mathbf{F}\mathbf{L}_{p}\mathbf{F}^{T}$ are convex. Hence, the objective function in Eq.(\ref{New13}) is convex. According to \cite{F-solve1}, we can easily obtain that the objective function in Eq.(\ref{New13}) monotonically decrease by using the updating rules in Algorithm~\ref{NewAlgorithm2}. Additionally, the objective functions in Eqs. (\ref{New15}) and (\ref{New20}) are convex w.r.t.  $\mathbf{S}$ and $\mathbf{P}$, and possess closed-form solutions, ensuring the convergence of the updates for $\mathbf{S}$ and $\mathbf{P}$, respectively. Therefore, we can conclude that the objective function in Eq.(\ref{New5}) monotonically decreases with each iteration of Algorithm~\ref{NewAlgorithm2}. Moreover, the convergence behavior of Algorithm~\ref{NewAlgorithm2} is further demonstrated through  experiments.  

\subsection{Complexity Analysis}
In Algorithm~\ref{NewAlgorithm2}, four variables are updated alternately by solving  Eqs.~(\ref{New6}), (\ref{New12}), (\ref{New15}), and (\ref{New20}). The optimization of Eq.~(\ref{New6}) is further transformed into solve Eq.~(\ref{New9}). According to \cite{JinHuang2014}, the time complexity for updating $\mathbf{W}$ in Eq.~(\ref{New9}) is $\mathcal{O}(d^{3})$. For updating $\mathbf{F}$, a proposed algorithm in \cite{F-solve1} is applied into solve Eq.~(\ref{New14}). Based on \cite{F-solve1}, we can calculate the computational complexity for updating $\mathbf{F}$, which is $\mathcal{O}(n^{2}c+c^{2}n)$. Eq.~(\ref{New15}) is solved by updating $\mathbf{S}$ row by row. According to the closed form solution of $\mathbf{S}$ in Eq.~(\ref{New19}), the time complexity of updating $\mathbf{S}$ is $\mathcal{O}(nk_{s}d)$. The optimization $\mathbf{P}$ in Eq.~(\ref{New20}) is similar to solving $\mathbf{S}$. In the same way, we can determine the time complexity of $\mathbf{P}$ as $\mathcal{O}(ck_{p}n)$. Therefore, the total computational complexity of Algorithm~\ref{NewAlgorithm2} is $\mathcal{O}((d^{3}+n^{2}c+c^{2}n+nk_{s}d+ck_{p}n)T)$, where $T$ is the number of iteration.

\section{Experiments}\label{sec:Experimental}
In this section, we compare Access-MFS with several state-of-the-art semi-supervised multi-label feature selection methods. We conduct extensive experiments on eight real-world benchmark datasets to demonstrate the effectiveness of our approach.

\subsection{Experimental Schemes}

\subsubsection{Datasets}
In our experiments, we collected eight real-world benchmark multi-label datasets, each with distinct statistical characteristics, to present the performance of various feature selection methods. These datasets include a biology dataset VirusGO\footnote{http://www.uco.es/kdis/mllresources\label{Source1}}, a medicine dataset  Medical\textsuperscript{\ref{Source1}}, and six text datasets: Langlog\textsuperscript{\ref{Source1}}, Enron\textsuperscript{\ref{Source1}}, Slashdot\textsuperscript{\ref{Source1}}, Recreation\footnote{http://www.lamda.nju.edu.cn/code\_MDDM.ashx\label{Source2}}, Computer\textsuperscript{\ref{Source2}} and, Science\textsuperscript{\ref{Source2}}. The detailed statistics for these multi-label datasets are summarized in Table~\ref{Table1}. Within this table, Label card. refers to the average number of labels per instance, while Label dens. reflects the label density, calculated as Label card. divided by the total number of labels.

\begin{table}[!htbp]
	\centering
	\small
	\tabcolsep 0pt
	\caption{A detail description of datasets.}  \label{Table1}
	\vspace*{-10pt}
	\renewcommand\tabcolsep{1.5pt} 
	\begin{flushleft}
		\def\temptablewidth{\textwidth}
			\begin{tabular*}{0.5\temptablewidth}{@{\extracolsep{\fill}}lccccc}
				\toprule
				Datasets & Instances & Features & Labels & Label card. & Label dens.\\
				\midrule[1pt]
				VirusGO		&  207		& 749	& 6	&1.2174  &0.2029	\\
				Medical		&  978		& 1449	& 45	&1.2454	&0.0277	\\
				Langlog		&  1460		& 1004	& 75	&1.1801	&0.0157	\\
				Enron		&  1702		& 1001	& 53	&3.3784	&0.0637	\\
				Slashdot	&  3782		& 1079	& 22	&1.1809	&0.0537	\\
				Recreation	&  5000		& 606	& 22	&1.4232	&0.0647	\\
				Computer	&  5000		& 681	& 33	&1.5082	&0.0457	\\
				Science	    &  5000		& 743	& 40	&1.4500  &0.0360	\\
				\hline
			\end{tabular*}
		\end{flushleft}
	\end{table}

	\subsubsection{Comparison Methods}
	To evaluate the performance of the proposed method Access-MFS, we compare it with several state-of-the-art multi-label feature selection methods. These include twelve semi-supervised methods (namely SMILE, FSSRDM, SCFS, LEDA, LSMR, SMLFS, SFAM, SMDR-IC, SFS-BLL, SMDR-MRC, SMFS and SFGR), one supervised method (LFFS), and one unsupervised method (UAFS-BH). The following is a brief introduction to the methods being compared.
	
	$\bullet$ \textbf{All-Fea} utilizes all original features for comparison.
	
	$\bullet$ \textbf{SMILE}~\cite{SMILE2017} uses label correlations to infer labels for unlabeled samples and embeds them into an adapted neighborhood graph model for feature selection.
	
	$\bullet$ \textbf{FSSRDM}~\cite{FSSRDM2018} selects features through an $\ell_{2,1}$-norm-based sparse regression model that incorporates maximizing the dependence between labels and features.
	
	$\bullet$ \textbf{SCFS}~\cite{SCFS2018} ensures consistency between feature and label spaces by leveraging spectral analysis to learn sample similarity during feature selection.
	
	$\bullet$ \textbf{LEDA}~\cite{LEDA2019} integrates a multi-label propagation mechanism into an extended linear discriminant analysis for selecting informative features. 
	
	$\bullet$ \textbf{LSMR}~\cite{LSMR2020} constructs a multi-label semi-supervised regression model for feature selection.
	
	$\bullet$ \textbf{SMLFS}~\cite{SMLFS2021} incorporates the feature graph and label graph into a logistic regression model with $\ell_{2,p}$ regularization to learn the regression coefficient matrix.
	
	$\bullet$ \textbf{SFAM}~\cite{SFAM2021} combines adaptive graph structure learning and local manifold learning in a unified feature selection framework.
	
	$\bullet$ \textbf{SMDR-IC}~\cite{SMDRIC2023} selects important features by incorporating label and sample correlations into a sparse regression model equipped with $\ell_{1}$-norm and $\ell_{2,1}$-norm regularization.
	
	$\bullet$ \textbf{SFS-BLL}~\cite{SFS-BLL2023} utilizes binary hash learning to generate pseudo labels and integrates them into a self-weighted sparse regression model for feature selection.
	
	$\bullet$ \textbf{SMDR-MRC}~\cite{SMDR-MRC2024} incorporates both $\ell_{1}$-norm and $\ell_{2,1}$-norm to extract specific and common features, respectively, and introduces a non-zero correlation constraint term to minimize redundancy and overlapping interference.
	
	$\bullet$ \textbf{SMFS}~\cite{SMFS2025} proposes a novel shared subspace learning framework that decomposes the feature selection matrix into a label-specific weight matrix and a shared subspace projection term, which explicitly captures the correlations among labels.
	
	$\bullet$ \textbf{SFGR}~\cite{SFGR2025} introduces a group decision-making mechanism into feature selection by using the Copeland function to fuse the ranking results of multiple feature subsets, thereby enhancing the stability and robustness of the selection process.
	
	$\bullet$ \textbf{LFFS}~\cite{LFFS2022} incorporates both global and local label correlation structural information into a sparse regression model to learn the feature selection matrix. 
	
	$\bullet$ \textbf{UAFS-BH}~\cite{UAFSBH2023} integrates binary hash constraints into the spectral embedding model to facilitate the learning of weakly-supervised labels and the selection of features.
	
	\subsubsection{Implementation Details}
	Several parameters need to be set in Access-MFS and the other compared methods. For Access-MFS, the regularization parameters $\lambda$, $\theta$ and $\mu$ are tuned using a grid search within the range $\{10^{-3}, 10^{-2}, 10^{-1}, 1, 10\}$. For all compared methods, the parameters are tuned using the same grid search strategy described in the original papers to achieve optimal results. For each dataset, following the approach of existing semi-supervised feature selection methods~\cite{SFS-BLL2023, SemiSetting}, we randomly select $\{10\%, 20\%, 30\%, 40\%, 50\%\}$ of the instances as the training set from the entire dataset, using the remainder as the testing set. Additionally, the test data also serve as unlabeled data for the semi-supervised multi-label feature selection methods. As determining the optimal number of selected features is still an open problem~\cite{Li2017}, we set the range for the number of selected features between 100 and 200 with an increment of 10 across all datasets. Then, Multi-Label K-Nearest Neighbor (MLKNN), as proposed in~\cite{MLKNN}, is executed on the selected features. Four commonly used evaluation metrics Average Precision~(AP)~\cite{ZhangandZhou2014}, Macro-F1~(MaF)~\cite{ZhangandZhou2014}, Ranking Loss~(RL)~\cite{ZhangandZhou2014} and One Error~(OE)~\cite{ZhangandZhou2014} are employed to assess the performance of different methods. For the Average Precision and Macro-F1 metrics, a higher value indicates better performance, while for Ranking Loss and One Error, a lower value indicates better performance. All methods are  implemented in MATLAB R2022b and executed on a desktop with an Intel Corei9-10900, CPU 2.80 GHz and 64 GB RAM. The experiments are independently conducted five times and the average results are reported for comparison. 
	
	\begin{table*}[!htbp]
		\caption{Performance comparison between Access-MFS and other methods on eight datasets in terms of AP and MaF.} \label{Table2}
		\vspace*{-10pt}
		\resizebox{\textwidth}{!}{
			\begin{tabu}{ccccccccccccccccc}
				\toprule[1pt]
				\multirow{2}{*}{Datasets} &\multicolumn{2}{c}{VirusGO} 	&\multicolumn{2}{c}{Medical}	&\multicolumn{2}{c}{Langlog}	&\multicolumn{2}{c}{Enron}	&\multicolumn{2}{c}{Slashdot}	&\multicolumn{2}{c}{Recreation}	&\multicolumn{2}{c}{Computer}	&\multicolumn{2}{c}{Science}\\ \tabucline[1pt]{2-17}
				&AP&MaF	  &AP&MaF	&AP&MaF	&AP&MaF	&AP&MaF	&AP&MaF	&AP&MaF	&AP&MaF \\ 
				\midrule[1pt]
				Access-MFS		&$\mathbf{.8918}$&$\mathbf{.5051}$		&$\mathbf{.8365}$&$\mathbf{.1951}$	&$\mathbf{.2914}$&$\mathbf{.0092}$	&$\mathbf{.6198}$&$\mathbf{.0737}$		&$\mathbf{.5859}$&$\mathbf{.1756}$		&$\mathbf{.5377}$&$\mathbf{.1455}$		&$\mathbf{.6536}$&$\mathbf{.1152}$	&$\mathbf{.5076}$&$\mathbf{.0746}$	\\
				All-Fea		&$.8621$&$.4577$		&$.7666$&$.1366$		&$.1846$&$.0011$		&$.6020$&$.0645$		&$.5055$&$.0455$		&$.4364$&$.0457$		&$.6278$&$.0587$		&$.4963$&$.0649$	\\
				SMILE		&$.6798$&$.1685$		&$.3985$&$.0031$		&$.1753$&$.0016$		&$.5692$&$.0468$		&$.4080$&$.0296$		&$.4293$&$.0344$		&$.6342$&$.0684$		&$.4686$&$.0442$	\\
				FSSRDM	    &$.6955$&$.2555$		&$.4107$&$.0067$		&$.2544$&$.0010$		&$.5947$&$.0633$		&$.4136$&$.0346$		&$.4320$&$.0451$		&$.6285$&$.0493$		&$.4220$&$.0128$	\\
				SCFS	    &$.8639$&$.4163$		&$.4249$&$.0070$		&$.2543$&$.0007$		&$.5674$&$.0486$		&$.4320$&$.0341$		&$.4518$&$.0502$		&$.6274$&$.0635$		&$.4367$&$.0261$	\\
				LEDA	    &$.6798$&$.1685$		&$.4074$&$.0032$		&$.1796$&$.0010$		&$.5989$&$.0616$		&$.4176$&$.0384$		&$.4331$&$.0437$		&$.6261$&$.0525$		&$.4301$&$.0130$	\\
				LSMR	    &$.6901$&$.2595$		&$.4937$&$.0300$		&$.2603$&$.0008$		&$.5987$&$.0659$		&$.4345$&$.0394$		&$.4396$&$.0401$		&$.6266$&$.0553$		&$.4242$&$.0142$	\\     
				SMLFS	    &$.8412$&$.3854$	&$.5318$&$.0489$		&$.2733$&$.0036$		&$.5946$&$.0606$		&$.4117$&$.0320$		&$.4250$&$.0297$		&$.6240$&$.0487$		&$.4318$&$.0201$	\\
				SFAM		&$.8543$&$.4341$		&$.5976$&$.0747$		&$.2279$&$.0012$		&$.5877$&$.0542$		&$.5037$&$.0802$		&$.4327$&$.0379$		&$.6312$&$.0571$		&$.4105$&$.0092$	\\
				SMDR-IC		&$.7768$&$.3375$		&$.4114$&$.0032$		&$.2486$&$.0023$		&$.5410$&$.0423$		&$.4076$&$.0243$		&$.4235$&$.0198$		&$.6201$&$.0522$		&$.4009$&$.0060$	\\
				SFS-BLL		&$.8442$&$.4088$		&$.7384$&$.1230$		&$.1693$&$.0012$		&$.6006$&$.0626$		&$.4154$&$.0333$		&$.4369$&$.0422$		&$.6270$&$.0585$		&$.4314$&$.0138$	\\
				SMDR-MRC		&$.8475$&$.4028$		&$.7969$&$.1873$		&$.2472$&$.0008$		&$.5980$&$.0620$		&$.4514$&$.0878$		&$.5111$&$.1209$		&$.6371$&$.0823$		&$.4693$&$.0516$	\\
				SMFS		&$.8612$&$.4823$		&$.7785$&$.1611$		&$.2635$&$.0011$		&$.5504$&$.0477$		&$.5467$&$.1316$		&$.5117$&$.1181$		&$.6336$&$.0680$		&$.4618$&$.0420$	\\
				SFGR		&$.8555$&$.3896$		&$.7727$&$.1342$		&$.2603$&$.0041$		&$.6042$&$.0601$		&$.4346$&$.0391$		&$.4109$&$.0190$		&$.6306$&$.0556$		&$.4245$&$.0177$	\\
				LFFS		&$.7884$&$.3594$		&$.5139$&$.0354$		&$.2538$&$.0010$		&$.5947$&$.0633$		&$.4390$&$.0439$		&$.4320$&$.0451$		&$.6285$&$.0493$		&$.4309$&$.0159$	\\
				UAFS-BH		&$.8636$&$.4273$		&$.4664$&$.0233$		&$.2662$&$.0040$		&$.5521$&$.0452$		&$.4010$&$.0301$		&$.4782$&$.0805$		&$.6319$&$.0604$		&$.4669$&$.0426$	\\
				\bottomrule[1pt]
		\end{tabu}}
	\end{table*}

	\begin{table*}[!htbp]
		\caption{Performance comparison between Access-MFS and other methods on eight datasets in terms of RL and OE.} \label{Table3}
		\centering
		\vspace*{-10pt}
		\resizebox{\textwidth}{!}{
			\begin{tabu}{ccccccccccccccccc}
				\toprule[1pt]
				\multirow{2}{*}{Datasets} &\multicolumn{2}{c}{VirusGO} 	&\multicolumn{2}{c}{Medical}	&\multicolumn{2}{c}{Langlog}	&\multicolumn{2}{c}{Enron}	&\multicolumn{2}{c}{Slashdot}	&\multicolumn{2}{c}{Recreation}	&\multicolumn{2}{c}{Computer}	&\multicolumn{2}{c}{Science}\\ \tabucline[1pt]{2-17}
				&RL&OE	  &RL&OE	&RL&OE	&RL&OE	&RL&OE	&RL&OE	&RL&OE	&RL&OE \\ 
				\midrule[1pt]
				Access-MFS		&$\mathbf{.0711}$&$\mathbf{.1677}$		&$\mathbf{.0449}$&$\mathbf{.2007}$	&$\mathbf{.1721}$&$\mathbf{.8395}$	&$\mathbf{.0968}$&$\mathbf{.3203}$		&$\mathbf{.1337}$&$\mathbf{.5354}$		&$\mathbf{.1655}$&$\mathbf{.5900}$		&$\mathbf{.0829}$&$\mathbf{.4281}$	&$\mathbf{.1248}$&$\mathbf{.6127}$\\
				All-Fea		&$.1008$&$.2065$		&$.0517$&$.3031$		&$.2050$&$.9363$		&$.1016$&$.3365$		&$.1685$&$.6401$		&$.1986$&$.7267$		&$.0919$&$.4507$		&$.1268$&$.6297$	\\
				SMILE		&$.2053$&$.5226$		&$.1416$&$.7335$		&$.2103$&$.9416$		&$.1085$&$.3894$		&$.1985$&$.7716$		&$.1990$&$.7405$		&$.0903$&$.4429$		&$.1367$&$.6573$	\\
				FSSRDM	    &$.1858$&$.5048$		&$.1390$&$.7140$		&$.1855$&$.8822$		&$.1076$&$.3393$		&$.1962$&$.7662$		&$.1973$&$.7653$		&$.09114$&$.4537$		&$.1501$&$.7155$	\\
				SCFS	    &$.0951$&$.2161$		&$.1387$&$.6932$		&$.1859$&$.8781$		&$.1077$&$.3957$		&$.1978$&$.7365$		&$.1944$&$.7070$		&$.0928$&$.4509$		&$.1447$&$.7046$	\\
				LEDA	    &$.2053$&$.5226$		&$.1401$&$.7242$		&$.2044$&$.9461$		&$.1065$&$.3312$		&$.1944$&$.7610$		&$.1951$&$.7397$		&$.0915$&$.4559$		&$.1488$&$.7030$	\\
				LSMR	    &$.1966$&$.5048$		&$.1139$&$.6317$		&$.1857$&$.8715$		&$.1070$&$.3318$		&$.1910$&$.7366$		&$.1962$&$.7263$		&$.0922$&$.4533$		&$.1473$&$.7167$	\\     
				SMLFS	    &$.1050$&$.2516$		&$.1102$&$.5792$		&$.1805$&$.8585$		&$.1044$&$.3220$		&$.2012$&$.7662$		&$.2023$&$.7417$		&$.0943$&$.4559$		&$.1456$&$.7123$	\\
				SFAM		&$.1069$&$.2258$		&$.0937$&$.5031$		&$.1945$&$.9027$		&$.1064$&$.3673$		&$.1710$&$.6356$		&$.1995$&$.7332$		&$.0917$&$.4451$		&$.1503$&$.7414$	\\
				SMDR-IC		&$.1317$&$.3839$		&$.1382$&$.7109$		&$.1781$&$.8884$		&$.1104$&$.3984$		&$.2054$&$.7667$		&$.2033$&$.7447$		&$.0955$&$.4581$		&$.1541$&$.7501$	\\
				SFS-BLL		&$.0951$&$.1952$		&$.0643$&$.3249$		&$.2044$&$.9457$		&$.1021$&$.3105$		&$.1959$&$.7632$		&$.1955$&$.7319$		&$.0901$&$.4579$		&$.1477$&$.7042$	\\
				SMDR-MRC		&$.1062$&$.2339$		&$.0551$&$.2628$		&$.1884$&$.8843$		&$.1027$&$.3430$		&$.1711$&$.7231$		&$.1739$&$.6278$		&$.0880$&$.4472$		&$.1330$&$.6609$	\\
				SMFS		&$.0956$&$.2097$		&$.0519$&$.2741$		&$.1789$&$.8795$		&$.1110$&$.3975$		&$.1546$&$.5794$		&$.1773$&$.6237$		&$.0917$&$.4404$		&$.1366$&$.6669$	\\
				SFGR		&$.0900$&$.2258$		&$.0506$&$.3007$		&$.1785$&$.8779$		&$.1038$&$.3328$		&$.1939$&$.7328$		&$.2071$&$.7640$		&$.0905$&$.4473$		&$.1440$&$.7393$	\\
				LFFS		&$.1388$&$.3436$		&$.1099$&$.5669$		&$.1855$&$.8822$		&$.1076$&$.3393$		&$.1869$&$.7060$		&$.1973$&$.7353$		&$.0914$&$.4537$		&$.1477$&$.7051$	\\
				UAFS-BH		&$.0956$&$.2048$		&$.1256$&$.6444$		&$.1739$&$.8671$		&$.1085$&$.3963$		&$.2073$&$.7779$		&$.1862$&$.6717$		&$.0914$&$.4426$		&$.1361$&$.6574$	\\
				\bottomrule[1pt]
		\end{tabu}}
	\end{table*}

	\subsection{Experimental Results and Performance Analysis}
	In this section, we demonstrate the superior performance of the proposed method Access-MFS compared with other competing methods in terms of AP, MaF, RL and OE. Table~\ref{Table2} summarizes the experimental results of various methods on eight multi-label datasets in terms of AP and MaF metrics, with the best performance highlighted in bold. From Table~\ref{Table2}, we can see that the proposed method Access-MFS consistently outperforms other methods across  all datasets. As to VirusGO and Recreation  datasets, Access-MFS achieves an average improvement of over 9\% in both AP and MaF metrics compared to all other methods. On Medical dataset, Access-MFS gains over 26\% and 12\% average improvement in terms of AP and MaF, respectively. As to  Slashdot dataset, Access-MFS outperforms other methods with more than 12\%  improvement in average AP and MaF. For  Science dataset, Access-MFS achieves an average improvement of over 6\% in AP and 4\% in MaF, respectively. On the Computer dataset, Access-MFS yields average improvement exceeding 2.5\% in AP and 5.7\% in MaF. On Langlog and Enron datasets, Access-MFS achieves an average improvement of over 3.6\% in AP and still outperforms all other methods in MaF. Additionally, Table~\ref{Table3} presents the experimental results of various methods in terms of RL and OE. From this table, we observe that the proposed Access-MFS consistently outperforms all other methods across all datasets. Furthermore, compared to the baseline method All-Fea, Access-MFS is superior across all datasets in terms of AP, MaF, RL, and OE, demonstrating the effectiveness of Access-MFS. Access-MFS also shows better performance than the supervised method LEFS on all datasets, indicating the effectiveness of the proposed semi-supervised learning strategy in enhancing feature selection performance.
	
\begin{figure*}[!htpb]
		\centering
		\includegraphics[width=1\textwidth]{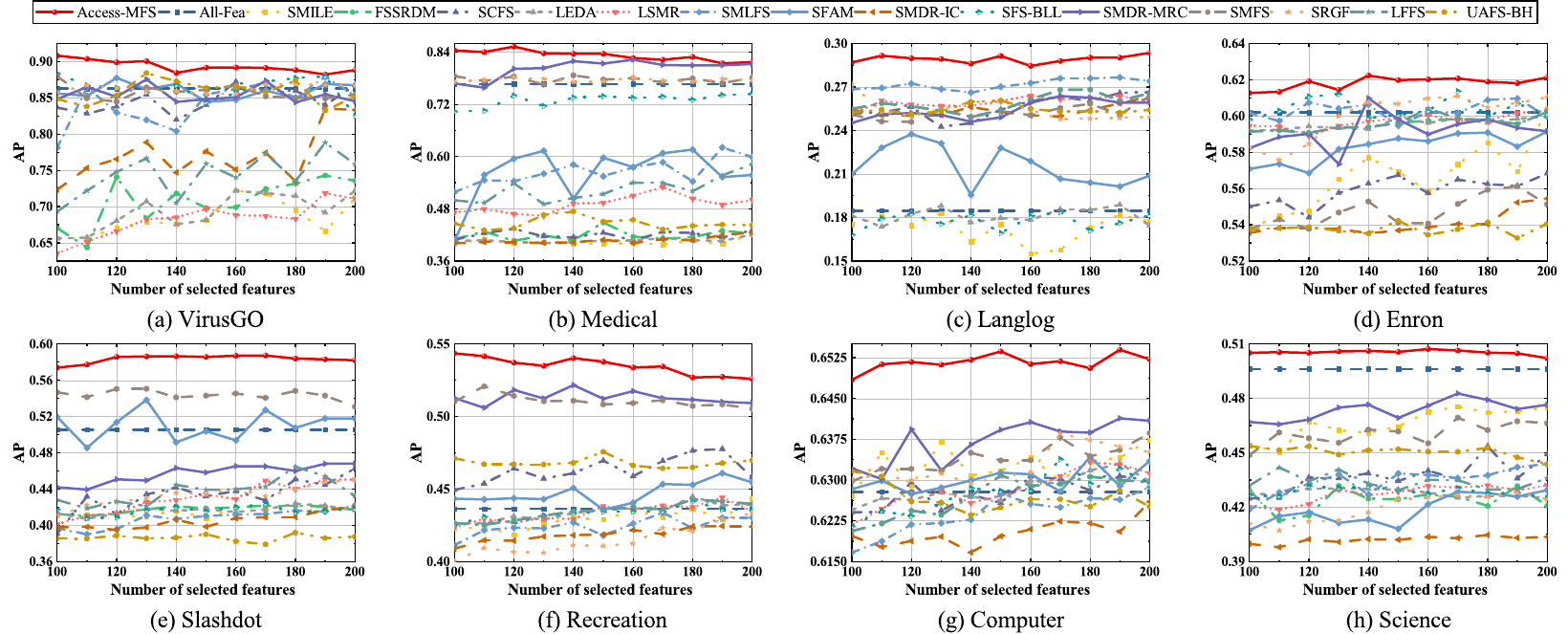}
		\caption{AP of different methods with varying numbers of selected features.}\small 
		\label{ApFeature}
	\end{figure*}

\begin{figure*}[!htpb]
		\centering
		\includegraphics[width=1\textwidth]{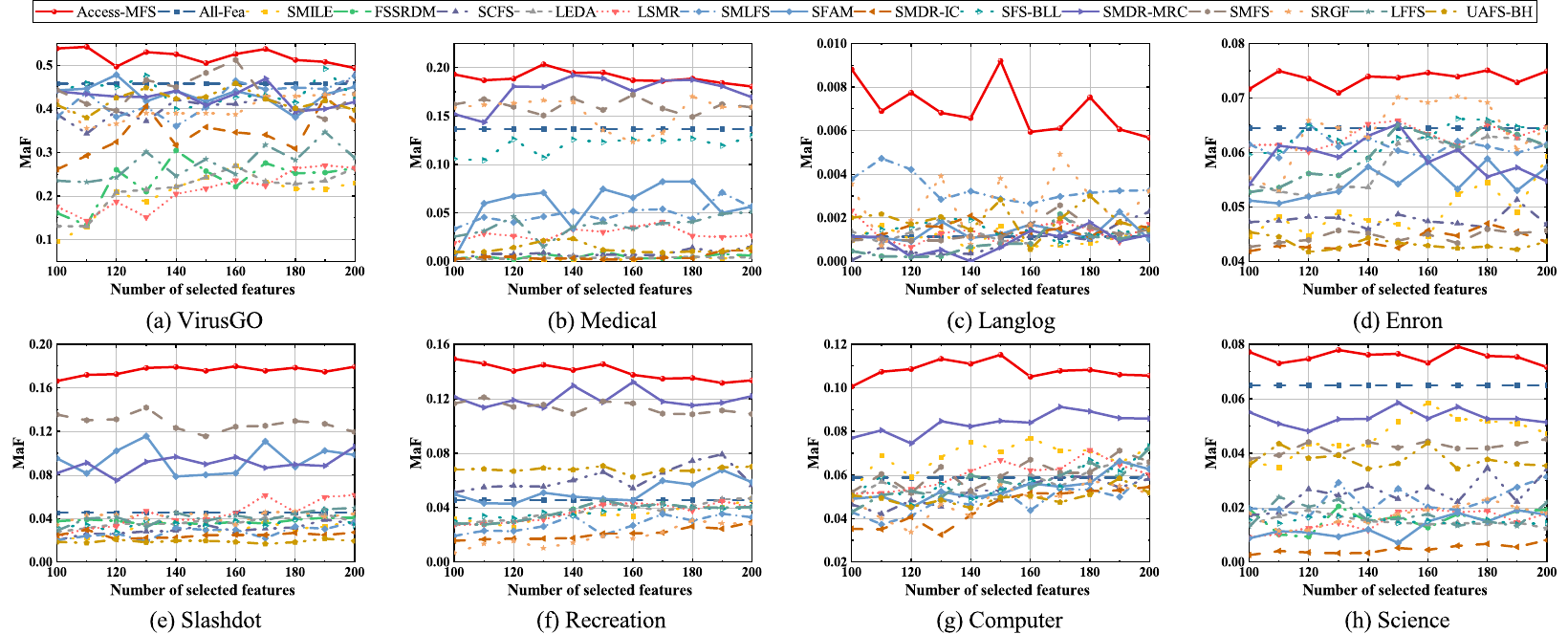}
		\caption{MaF of different methods with varying numbers of selected features.}\small 
		\label{MafFeature}
	\end{figure*}

\begin{figure*}[!htpb]
		\centering
		\includegraphics[width=1\textwidth]{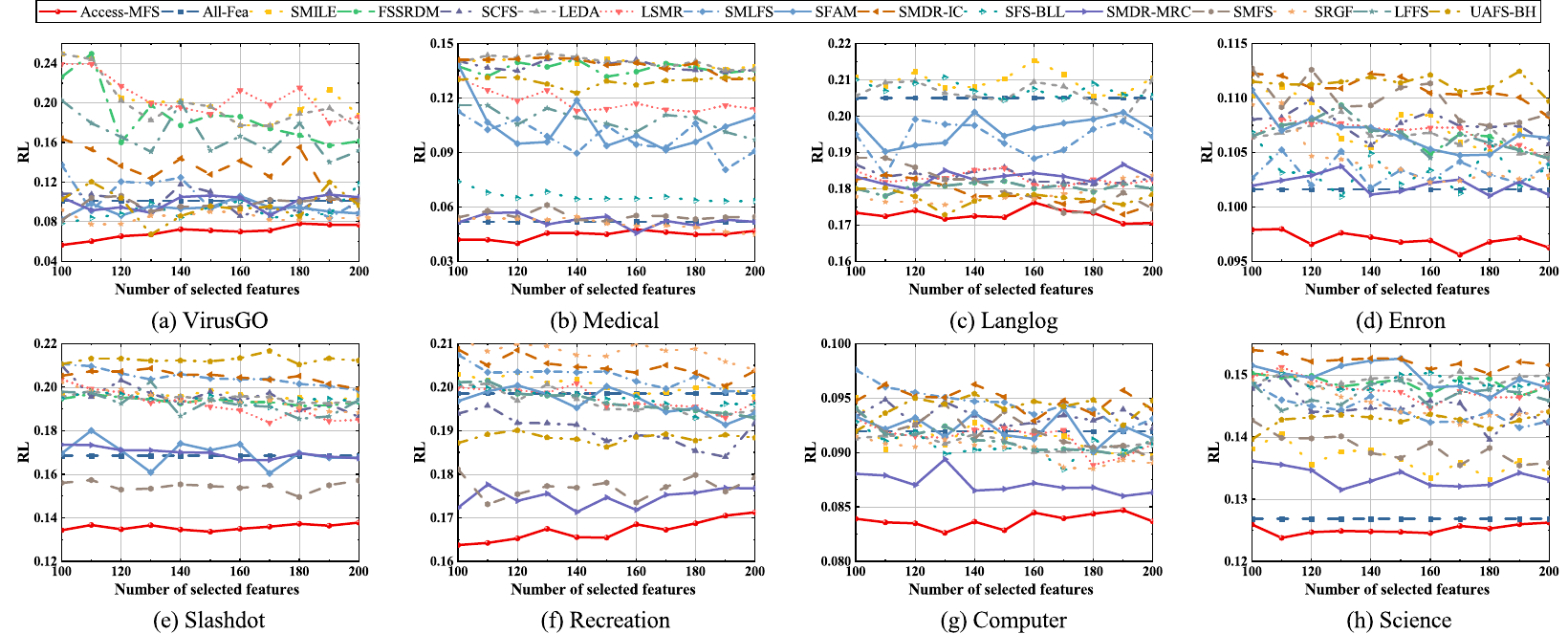}
		\caption{RL of different methods with varying numbers of selected features.}\small 
		\label{Rlfeature}
	\end{figure*}

\begin{figure*}[!htpb]
		\centering
		\includegraphics[width=1\textwidth]{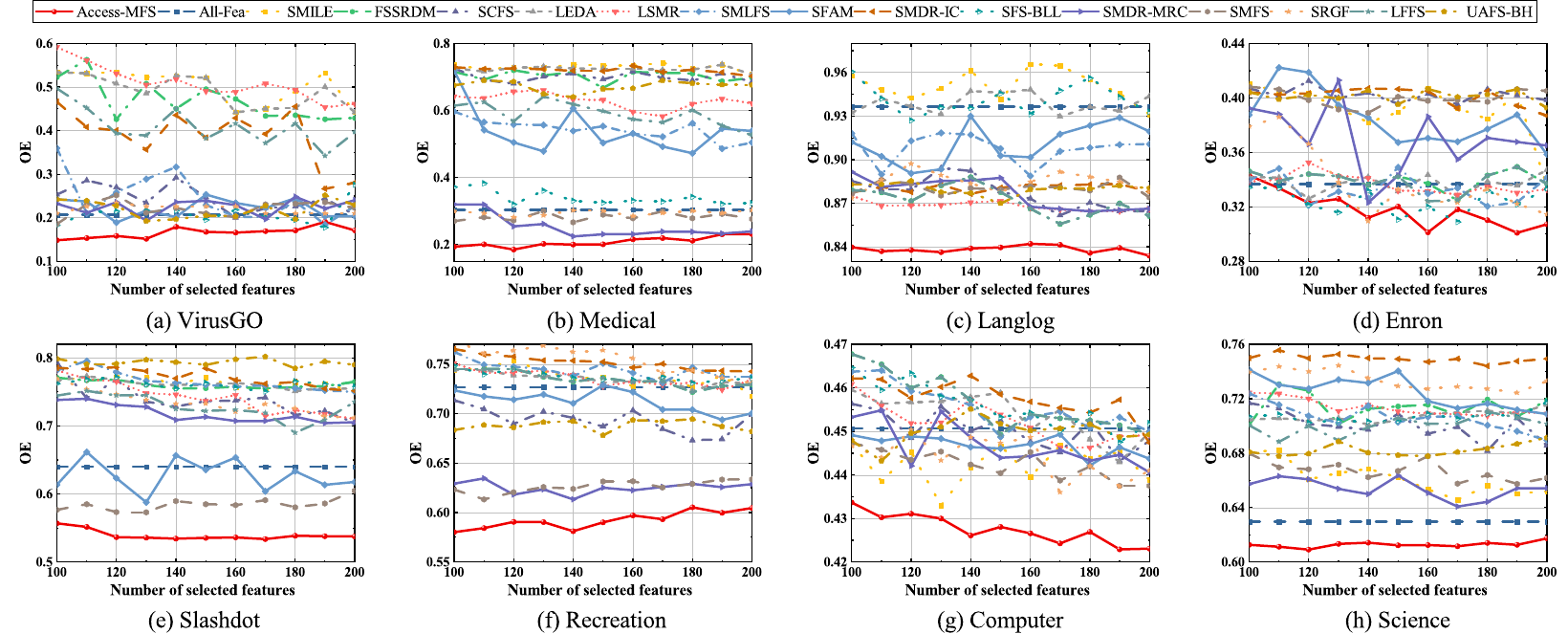}
		\caption{OE of different methods with varying numbers of selected features.}\small 
		\label{Oefeature}
	\end{figure*}

\begin{figure*}[!htpb]
		\centering
		\includegraphics[width=1\textwidth]{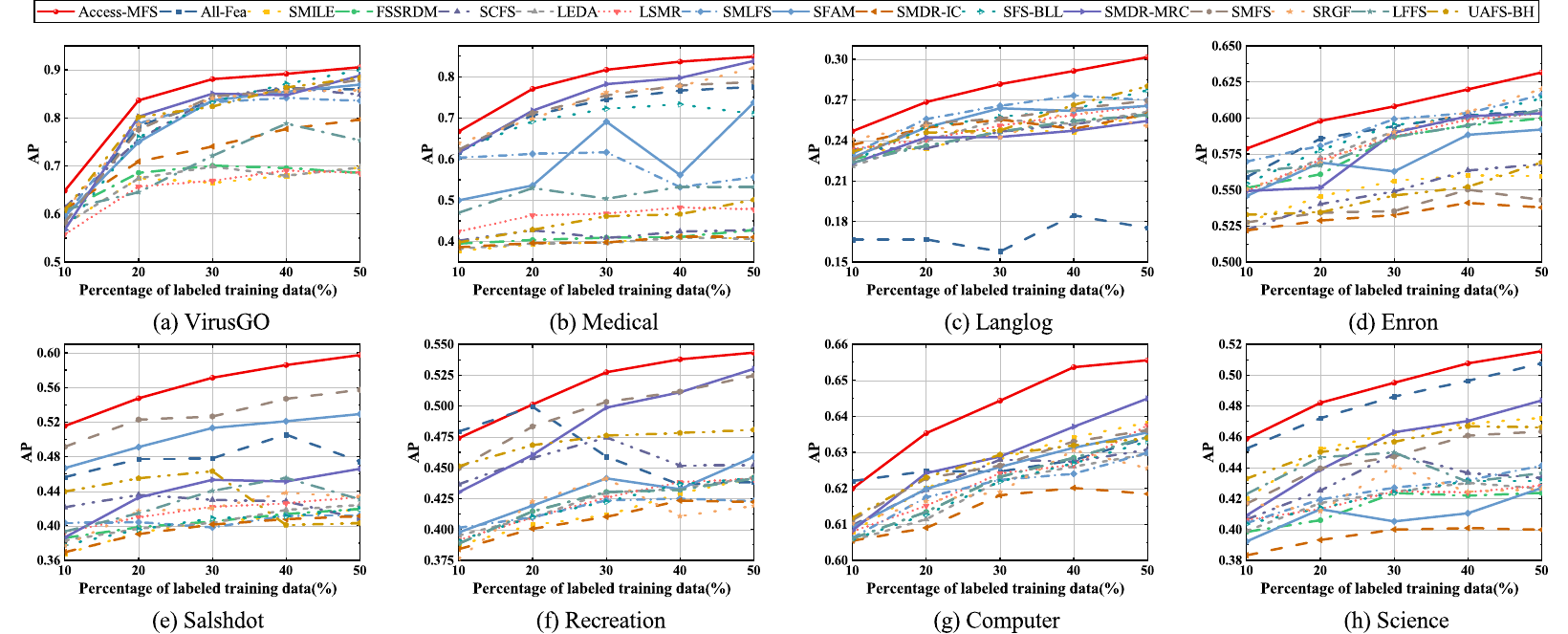}
		\caption{AP of different methods on eight datasets with varying percentages of labeled training data.}\small 
		\label{ApLabel}
	\end{figure*}

\begin{figure*}[!htpb]
		\centering
		\includegraphics[width=1\textwidth]{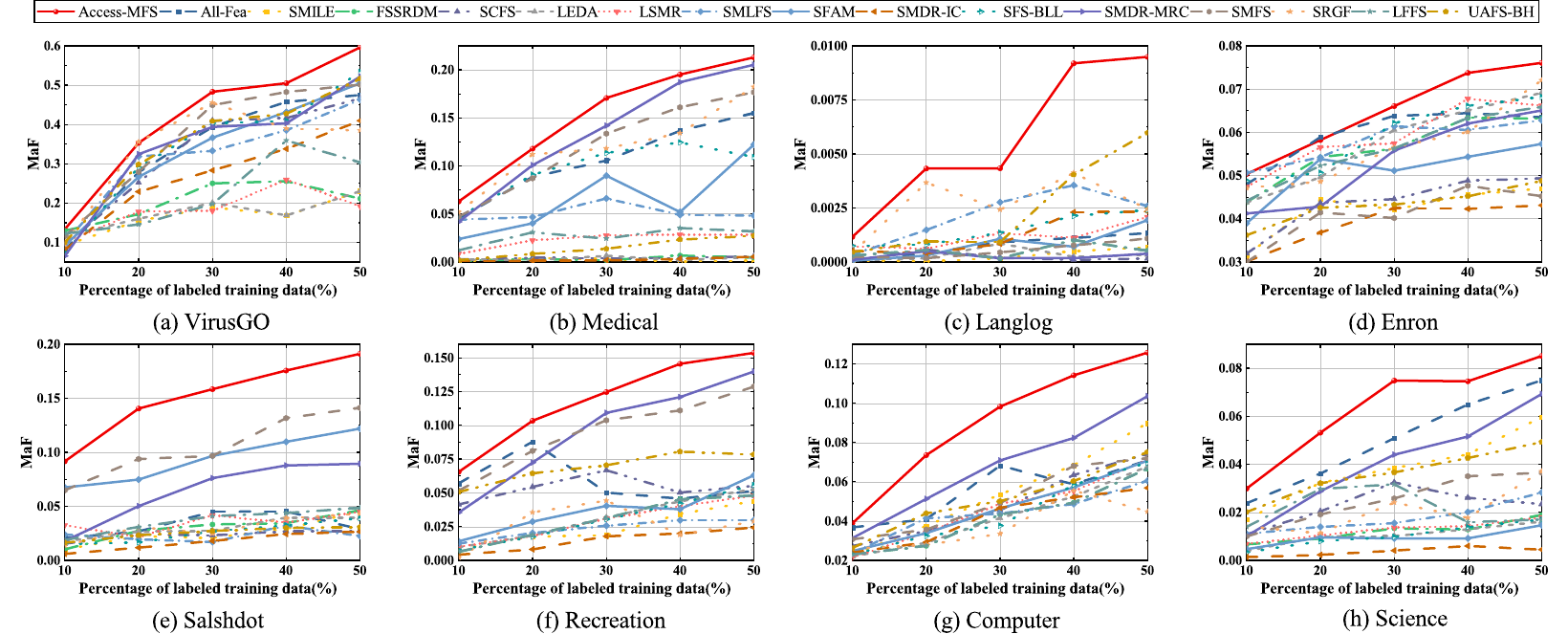}
		\caption{MaF of different methods on eight datasets with varying percentages of labeled training data.}\small 
		\label{MafLabel}
	\end{figure*}

\begin{figure*}[!htpb]
		\centering
		\includegraphics[width=1\textwidth]{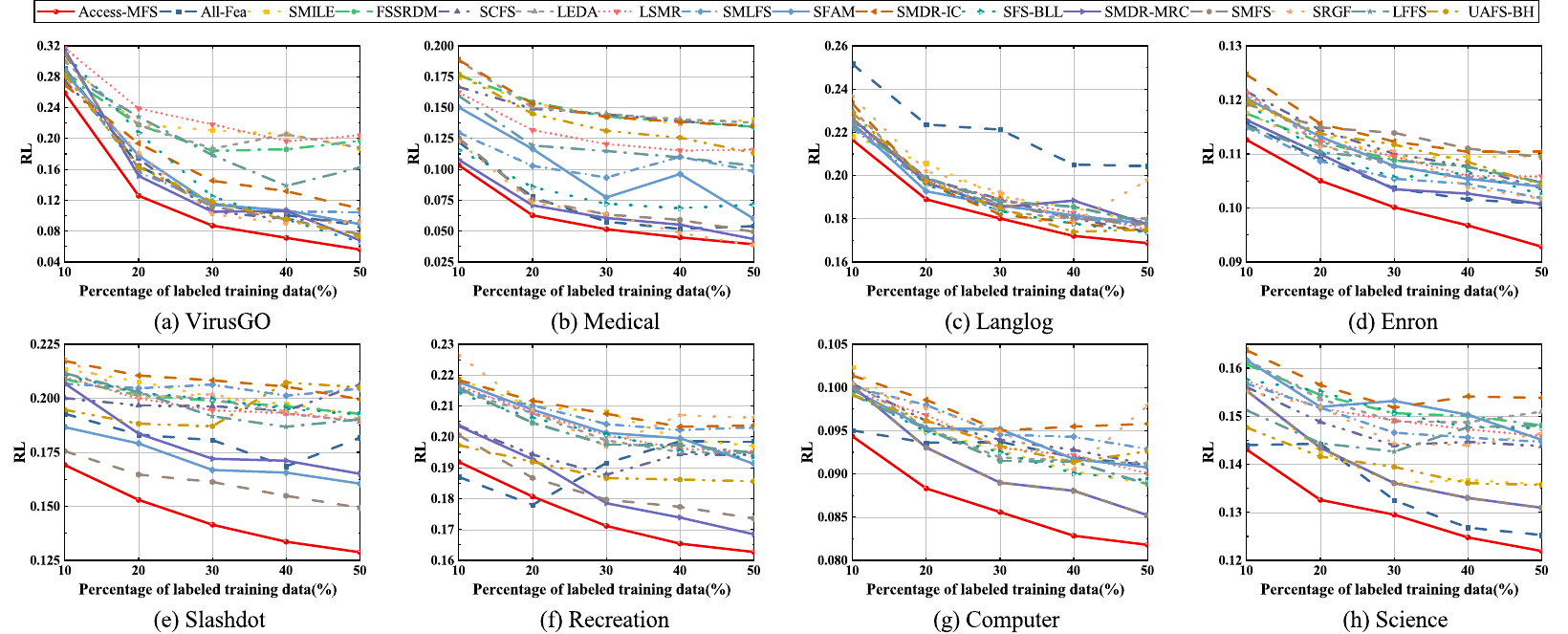}
		\caption{RL  of different methods on eight datasets with varying percentages of labeled training data.}\small 
		\label{RlLabel}
	\end{figure*}

\begin{figure*}[!htpb]
		\centering
		\includegraphics[width=1\textwidth]{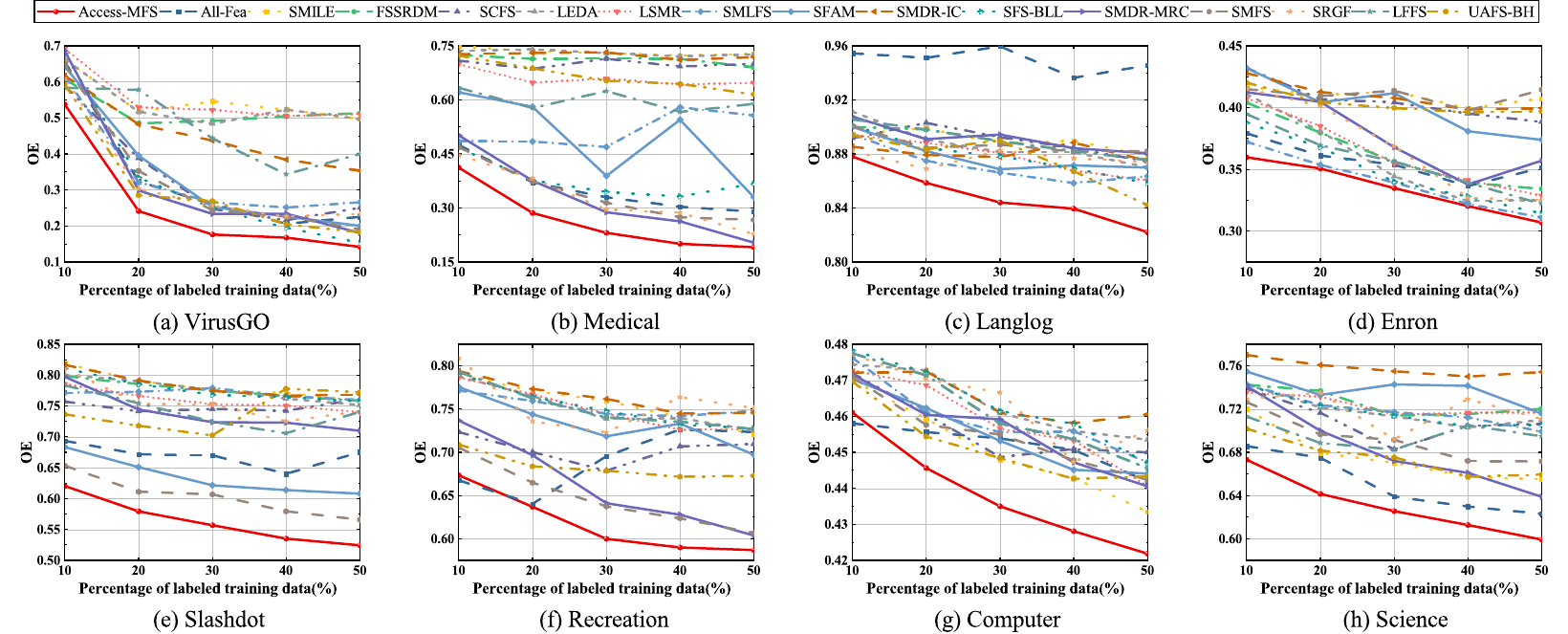}
		\caption{OE of different methods on eight datasets with varying percentages of labeled training data.}\small 
		\label{OeLabel}
	\end{figure*}

	Due to the difficulty in determining the optimal number of selected features, we also present the performance (AP, MaF, RL and OE) of different methods as the number of selected features varies.  Figs.~\ref{ApFeature}, \ref{MafFeature}, \ref{Rlfeature} and \ref{Oefeature} respectively show the AP, MaF, RL and OE values for different numbers of selected features, with the labeled instance ratio fixed at 40\%. As illustrated in these four figures, the proposed Access-MFS method outperforms other methods in most cases, with the number of selected features varying  from 100 to 200 on each dataset. Furthermore, we also investigate the performance of different methods with varying ratios of labeled instances. Figs.~\ref{ApLabel}, \ref{MafLabel}, \ref{RlLabel} and \ref{OeLabel} respectively show the AP, MaF, RL and OE of different methods when the ratio of labeled instances ranges from 10\% to 50\%. From these figures, we can see that our method achieves better performance in most cases. In addition, our method performs better as the ratio of labeled instances increases, indicating that our approach effectively leverages labeled information to guide the collaborative learning of  sample and label similarity graphs, ultimately enhancing feature selection performance. The superior performance of the proposed method is attributed to the adaptive collaborative learning of similarity graphs from both the sample and label perspectives, which are embedded into a generalized regression model with an extended uncorrelated constraint and a consistency constraint between predicted and existing labels.
	
	\subsection{Parameter Sensitivity Experiment}
\begin{figure*}[!htpb]
		\centering
		\includegraphics[width=1\textwidth]{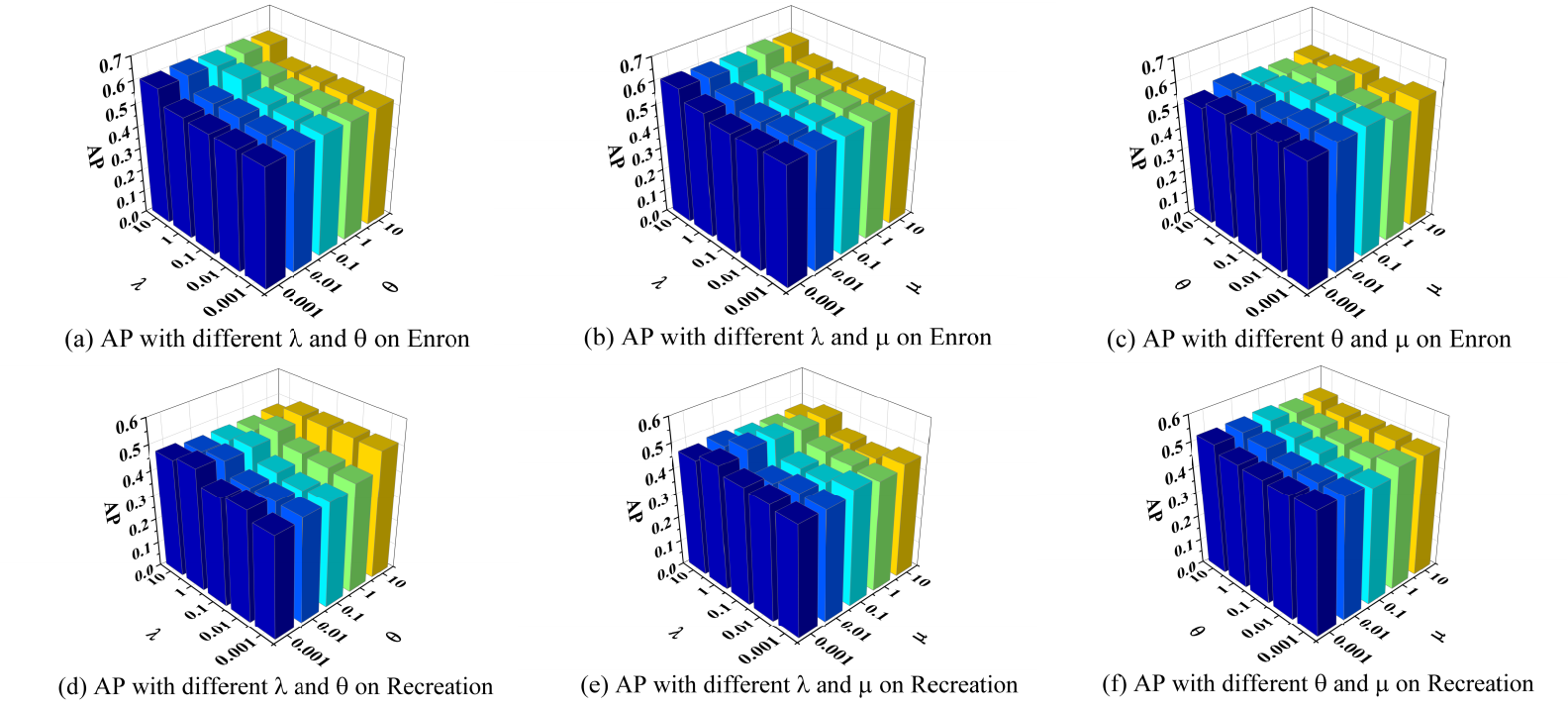}
		\caption{Variations of AP with different parameters on Enron and Recreation datasets}\small 
		\label{Fig9}
	\end{figure*}

	In this section, we conduct experiments to investigate the sensitivity of the proposed Access-MFS to three parameters, i.e., $\lambda$, $\theta$, and ${\mu}$. Fig.~\ref{Fig9} shows the performance variations w.r.t. AP on Enron and Recreation datasets  when two of the parameters $\lambda$, $\theta$, and ${\mu}$ are varied while the third is kept fixed. We can observe that the parameters  $\lambda$ and $\theta$ are relatively more sensitive compared to the parameter ${\mu}$. Besides, when ${\mu}$ is fixed, the proposed method achieves prominent performances with the parameter combinations of $\lambda=10$ or $1$ and $\theta=10$ in most cases. Hence, We can empirically fine-tune $\lambda$ and  $\theta$ within a narrower range discussed to achieve a better performance.
	
	\begin{figure*}[!htpb]
		\centering
		\includegraphics[width=1\textwidth]{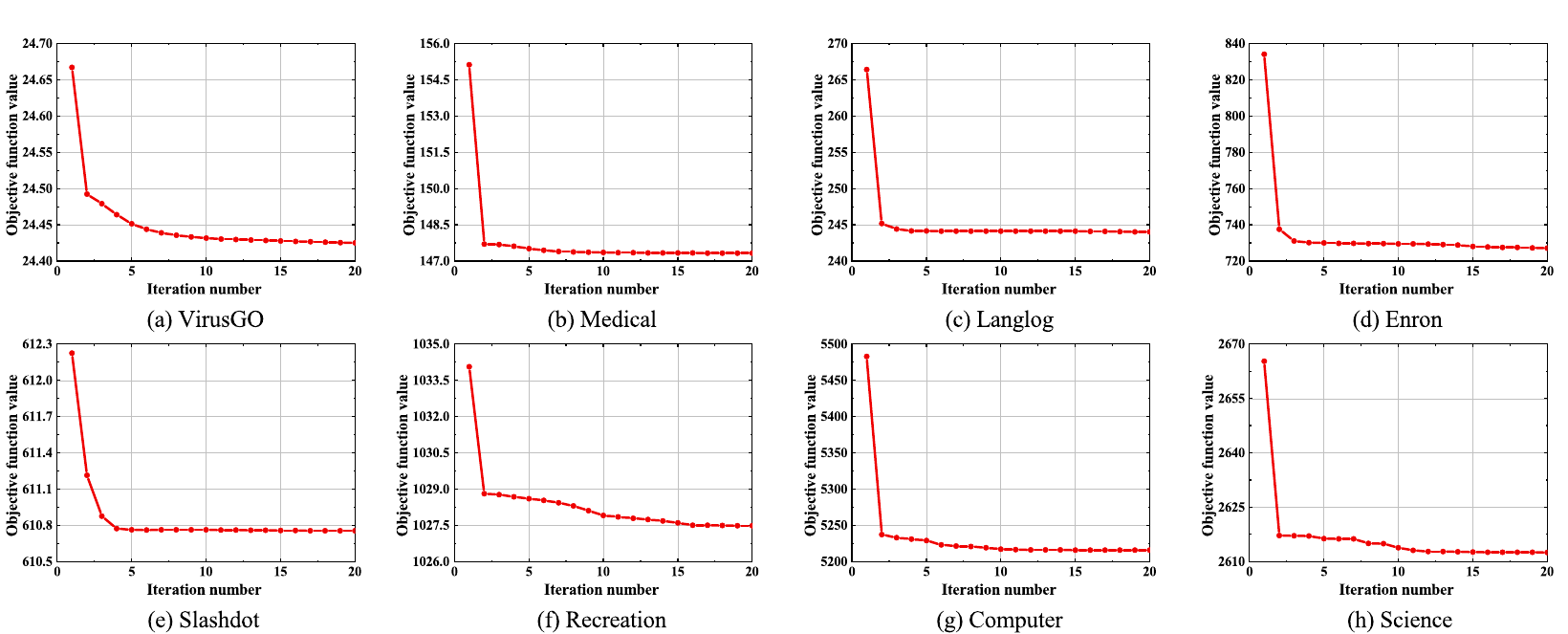}
		\caption{Convergence curves of Access-MFS.}\small 
		\label{Fig10}
	\end{figure*}

	\subsection{Convergence Experiment}
	The theoretical convergence of the proposed Access-MFS algorithm has been established in Section 5.1. This subsection provides experimental validation of its convergence and analyzes the speed of convergence. Fig.~\ref{Fig10} illustrates the convergence curves of Access-MFS across eight datasets, showing the variation in the objective function value over the course of iterations. As illustrated in Fig.~\ref{Fig10}, the convergence curves steeply fall within a few iterations and approach stability almost within 15 iterations. The experimental results demonstrate that the proposed method can converge effectively.
	
	\subsection{Ablation Study}
	In this section, we conduct an ablation study to evaluate the effects of the adaptive collaborative correlation learning module for the instance similarity graph and label similarity graph in the proposed method. To this end, three variant methods of Access-MFS are designed for comparison as follows:
	
	(1) Access-MFS-\uppercase\expandafter{\romannumeral1}:  It is only composed of the instance similarity graph learning module in Access-MFS. 
	\begin{equation}\label{NAbl1}
		\begin{aligned}
			&\min _{\Xi}\|\mathbf{X}^{T} \mathbf{W}+\mathbf{1}_n \mathbf{b}^{T}-\mathbf{F}\|_F^{2}+{ \|{\mathbf{F}_{l}-\mathbf{Y}_{l}}\|_F^2}+\lambda\|\mathbf{W}\|_{2,1}\\
			&\quad\quad+\theta (\operatorname{Tr}(\mathbf{W}^{T}\mathbf{X}\mathbf{L}_{s}\mathbf{X}^{T}\mathbf{W})
			+\operatorname{Tr}(\mathbf{F}^{T} \mathbf{L}_{s} \mathbf{F})+\alpha \|\mathbf{S}\|_F^2)\\
			&{s.t.}~\mathbf{W}^{T}\mathbf{R} \mathbf{W}=\mathbf{I}, s_{i i}=0, s_{i j} \geq 0, \mathbf{S}\mathbf{1}_{n}=\mathbf{1}_{n}.
		\end{aligned}
	\end{equation}
	
	(2) Access-MFS-\uppercase\expandafter{\romannumeral2}:  It is only composed of the label similarity graph learning module in Access-MFS. 
	\begin{equation}\label{NAbl2}
		\begin{aligned}
			&\min _{\Xi}\|\mathbf{X}^{T} \mathbf{W}+\mathbf{1}_n \mathbf{b}^{T}-\mathbf{F}\|_F^{2}+{ \|{\mathbf{F}_{l}-\mathbf{Y}_{l}}\|_F^2}+\lambda\|\mathbf{W}\|_{2,1}\\
			&\quad\quad+\mu(\operatorname{Tr}(\mathbf{F} \mathbf{L}_{p} \mathbf{F}^{T})+\beta \|\mathbf{P}\|_F^2)\\
			&{s.t.}~\mathbf{W}^{T}\mathbf{R} \mathbf{W}=\mathbf{I},p_{i i}=0, p_{i j} \geq 0,\mathbf{P}\mathbf{1}_{c}=\mathbf{1}_{c}.
		\end{aligned}
	\end{equation}
	
	(3) Access-MFS-\uppercase\expandafter{\romannumeral3}:  It performs feature selection without the adaptive collaborative learning of the instance similarity graph and label similarity graph modules in Access-MFS.
	\begin{equation}\label{NAbl3}
		\begin{aligned}
			&\min _{\Xi}\|\mathbf{X}^{T} \mathbf{W}+\mathbf{1}_n \mathbf{b}^{T}-\mathbf{F}\|_F^{2}+{ \|{\mathbf{F}_{l}-\mathbf{Y}_{l}}\|_F^2}+\lambda\|\mathbf{W}\|_{2,1}\\
			&{s.t.}~\mathbf{W}^{T}\mathbf{R} \mathbf{W}=\mathbf{I}.
		\end{aligned}
	\end{equation}
	
	By comparing the proposed Access-MFS with its three variant methods, Table~\ref{Table4} presents the ablation results for AP and MaF, while Table~\ref{Table5} shows the results for RL and OE. From those tables, we can see that our method outperforms methods that use only instance similarity graph learning or only label similarity graph learning (namely, Access-MFS-\uppercase\expandafter{\romannumeral1} and Access-MFS-\uppercase\expandafter{\romannumeral2}). Additionally, methods with only one type of similarity graph learning (Access-MFS-\uppercase\expandafter{\romannumeral1} and Access-MFS-\uppercase\expandafter{\romannumeral2}) perform better than those without any similarity graph learning, i.e., Access-MFS-\uppercase\expandafter{\romannumeral3}. This demonstrates the effectiveness of the proposed adaptive collaborative correlation learning module, which can automatically capture the local geometric structure of samples and accurately learn label correlations using existing labels. These two processes mutually enhance each other, improving feature selection performance.
	
	\begin{table*}[!htbp]
		\caption{Means of AP and MaF for different variants of Access-MFS on eight datasets.} \label{Table4}
		\vspace*{-10pt}
		\resizebox{\textwidth}{!}{
			\begin{tabu}{cccccccccc}
				\toprule[1pt]
				Metircs	&Methods	&VirusGO &Medical	&Langlog	&Enron	&Slashdot	&Recreation	&Computer &Science	\\ 
				\midrule[1pt]
				\multirow{4}{*}{AP}	&Access-MFS	&$\mathbf{.8918}$		&$\mathbf{.8365}$	&$\mathbf{.2914}$	&$\mathbf{.6198}$		&$\mathbf{.5859}$		&$\mathbf{.5377}$		&$\mathbf{.6536}$	&$\mathbf{.5027}$\\	
				&Accsess-MFS-\uppercase\expandafter{\romannumeral1}		&$.8624$		&$.8009$		&$.2797$		&$.5805$		&$.5453$		&$.5109$		&$.6477$		&$.4862$\\
				&Accsess-MFS-\uppercase\expandafter{\romannumeral2}	&$.8659$		&$.7967$		&$.2871$		&$.5854$		&$.5521$		&$.5100$		&$.6453$	&$.4879$		\\
				&Access-MFS-\uppercase\expandafter{\romannumeral3}	&$.8570$		&$.7753$		&$.2652$		&$.5399$		&$.4502$		&$.4841$		&$.6349$	&$.4469$		\\
				\tabucline[1pt]{1-10}
				\multirow{4}{*}{MaF}	&Access-MFS	&$\mathbf{.5051}$		&$\mathbf{.1951}$	&$\mathbf{.0092}$	&$\mathbf{.0737}$		&$\mathbf{.1756}$		&$\mathbf{.1455}$		&$\mathbf{.1152}$	&$\mathbf{.0746}$\\		
				&Accsess-MFS-\uppercase\expandafter{\romannumeral1}		&$.4328$		&$.1770$		&$.0048$		&$.0688$		&$.1290$		&$.1199$		&$.0890$		&$.0622$		\\
				&Accsess-MFS-\uppercase\expandafter{\romannumeral2}	&$.4515$		&$.1644$		&$.0027$		&$.0682$		&$.1346$		&$.1187$		&$.0956$	&$.0605$		\\
				&Access-MFS-\uppercase\expandafter{\romannumeral3}	&$.4276$		&$.1545$		&$.0009$		&$.0610$		&$.0878$		&$.0970$		&$.0803$	&$.0340$		\\
				\bottomrule[1pt]
		\end{tabu}}
	\end{table*}	
	
	\begin{table*}[!htbp]
		\caption{Means of RL and OE for different variants of Access-MFS on eight datasets.} \label{Table5}
		\vspace*{-10pt}
		\resizebox{\textwidth}{!}{
			\begin{tabu}{cccccccccc}
				\toprule[1pt]
				Metircs	&Methods	&VirusGO 	&Medical	&Langlog	&Enron	&Slashdot	&Recreation	&Computer	&Science\\ 
				\midrule[1pt]					
				\multirow{4}{*}{RL}	&Access-MFS	&$\mathbf{.0711}$		&$\mathbf{.0449}$	&$\mathbf{.1721}$	&$\mathbf{.0968}$		&$\mathbf{.1337}$		&$\mathbf{.1655}$		&$\mathbf{.0829}$	&$\mathbf{.1248}$\\	
				&Accsess-MFS-\uppercase\expandafter{\romannumeral1}		&$.0820$		&$.0510$		&$.1726$		&$.1076$		&$.1523$		&$.1750$		&$.0853$		&$.1294$		\\
				&Accsess-MFS-\uppercase\expandafter{\romannumeral2}	&$.0873$		&$.0500$		&$.1762$		&$.1074$		&$.1522$		&$.1743$		&$.0866$	&$.1295$		\\
				&Access-MFS-\uppercase\expandafter{\romannumeral3}	&$.0969$		&$.0563$		&$.1794$		&$.1115$		&$.1735$		&$.1793$		&$.0878$	&$.1379$		\\
				\tabucline[1pt]{1-10}
				\multirow{4}{*}{OE}	&Access-MFS	&$\mathbf{.1677}$	&$\mathbf{.2007}$	&$\mathbf{.8395}$	&$\mathbf{.3203}$		&$\mathbf{.5354}$		&$\mathbf{.5900}$		&$\mathbf{.4281}$	&$\mathbf{.6157}$	\\
				&Accsess-MFS-\uppercase\expandafter{\romannumeral1}		&$.2113$		&$.2515$		&$.8491$		&$.3317$		&$.5846$		&$.6263$		&$.4339$		&$.6413$		\\
				&Accsess-MFS-\uppercase\expandafter{\romannumeral2}	&$.1984$		&$.2590$		&$.8422$		&$.3389$		&$.5740$		&$.6285$		&$.4387$	&$.6358$		\\
				&Access-MFS-\uppercase\expandafter{\romannumeral3}	&$.2226$		&$.2891$		&$.8731$		&$.3408$		&$.7203$		&$.6661$		&$.4482$	&$.6917$		\\
				\bottomrule[1pt]
		\end{tabu}}
	\end{table*}	
	
	\section{Conclusion}\label{sec:Conclusions}
	To address the high dimensionality of multi-label data and the difficulty of obtaining labels for all instances, we proposed a novel semi-supervised multi-label feature selection method named Access-MFS, which uses only a subset of labeled samples. Unlike existing methods that use predefined approaches to describe either sample similarity or label similarity separately, our method enhanced the performance of semi-supervised multi-label feature selection by adaptively and collaboratively learning both sample and label similarity correlations. An generalized regression model equipped with an extended uncorrelated constraint was presented to select informative features from semi-supervised multi-label data. Meanwhile, the adaptive collaborative learning module of instance similarity graphs and label similarity graphs was integrated into the generalized regression model to preserve sample similarity in both high-dimensional and low-dimensional spaces, assign similar labels to similar samples, and establish strong label correlations for consistent predictions. An alternative iterative optimization algorithm was also developed to solve the proposed objective function. The experimental results on real-world semi-supervised multi-label data showed that Access-MFS outperformed the state-of-the-art methods.
	
	\section*{Acknowledgments}
	This work was supported by the National Natural Science Foundation of China (No. 72495122), the Natural Science Foundation Project of Sichuan Province (No. 2024NSFSC0504), and the Youth Fund Project of Humanities and Social Science Research of Ministry of Education (No. 21YJCZH045).


\begin{thebibliography}{1}
		
		\bibitem{ZhangandZhou2014}M.L. Zhang and Z.H. Zhou, "A review on multi-Label learning algorithms", \textit{IEEE Trans. Knowl. Data Eng.}, 26 (8) (2014) 1819-1837.
		
		\bibitem{supervisedmulti-label1}M.L. Zhang and L. Wu, "Lift: Multi-label learning with label-specific features," \textit{IEEE Trans. Pattern Anal. Mach. Intell.}, 37 (1) (2015) 107-120.
		\bibitem{VideoCla}W. Lu, D. Li, L. Nie, P. Jing and Y. Su, "Learning dual low-rank representation for multi-label micro-video classification," \textit{IEEE Trans. Multimedia}, 25 (2023) 77-89.
		
		\bibitem{TextCla}T. Jiang, D. Wang, L. Sun, H. Yang, Z. Zhao, and F. Zhuang, "LightXML: Transformer with dynamic negative sampling for high-performance extreme multi-label text classification," \textit{AAAI Conf. Artif. Intell.}, 35 (9) (2021) 7987-7994.
		
		\bibitem{ImageAnn}C. Wang, S.C Yan, L. Zhang and H. J. Zhang, "Multi-label sparse coding for automatic image annotation," \textit{IEEE Comput. Soc. Conf. Comput. Vis. Pattern Recognit.}, 2009, pp. 1643-1650.
		
		\bibitem{Qian2023}W. Qian, J. Huang, F. Xu, W. Shu and W. Ding, "A survey on multi-label feature selection from perspectives of label fusion," \textit{Inf. Fusion}, 100 (2023) 101948.
		
		\bibitem{Hu2022}L. Hu, L. Gao, Y. Li, P. Zhang and W. Gao, "Feature-specific mutual information variation for multi-label feature selection," \textit{Inf. Sci.}, 593 (2022) 449-471.
		
		\bibitem{Li2017}J. Li, K. Cheng, S. Wang, F. Morstatter, R. P. Trevino, J. Tang and H. Liu, "Feature selection: A data perspective," \textit{ACM Comput. Surv.}, 50 (6) (2017) 1-45.
		
		\bibitem{Lapscore}X. He, D. Cai and P. Niyogi, "Laplacian score for feature selection," \textit{Proc. Adv. Neural Inf. Process. Syst.}, 2005, pp. 507-514.
		
		\bibitem{Fu2021}Y. Fu, X. Liu, S. Sarkar and T. Wu, "Gaussian mixture model with feature selection: An embedded approach," \textit{Comput. Ind. Eng.}, 152 (2021) 107000.
		
		\bibitem{Zhao2007}Z. Zhao and L. Huan, "Spectral feature selection for supervised and unsupervised learning," \textit{Proc. 24th Int. Conf. Mach. Learn.}, 2007, pp. 1151-1157.
		
		\bibitem{Wu2010}F. Wu, Y. Yuan and Y. Zhuang, "Heterogeneous feature selection by group lasso with logistic regression," \textit{Proc. 18th ACM Int. Conf. Multimedia}, 2010, pp. 983-986.
		
		\bibitem{JianL2016}L. Jian, J. Li, K. Shu and H. Liu, "Multi-label informed feature selection," \textit{ Proc. Int. Joint Conf. Artif. Intell.}, 2016, pp. 1627-1633.
		
		\bibitem{Braytee2017}A. Braytee, W. Liu, D. R. Catchpoole and P. J. Kennedy, "Multi-label feature selection using correlation information," \textit{Proc. ACM Conf. Inf. Knowl. Management}, 2017, pp. 1649-1656.
		
		\bibitem{JZhang2019}J. Zhang, Z. Luo, C. Li, C. Zhou and S. Li, "Manifold regularized discriminative feature selection for multi-label learning," \textit{Pattern Recognit.}, 95 (2019) 136–150.
		
		\bibitem{LFFS2022}Y.L. Fan, B.H. Chen, W.Q. Huang, J.H. Liu, W. Weng and W.Y. Lan, "Multi-label feature selection based on label correlations and feature redundancy," \textit{Knowl. Based Syst.}, 241 (2022) 108256.
		
		\bibitem{JZhang2023}J. Zhang, H. Wu, M. Jiang, J. Liu, S. Li, Y. Tang and J. Long, "Group-preserving label-specific feature selection for multi-label learning," \textit{Expert Syst. Appl.}, 213 (2023) 118861.
		
		\bibitem{Ma2012}Z.Ma, F. Nie, Y. Yang, J. R. Uijlings and N. Sebe, "Web image annotation via subspace-sparsity collaborated feature selection," \textit{IEEE Trans. Multimedia}, 14 (4) (2012) 1021-1030.
		
		\bibitem{semi-supervised1}Z. Xu, I. King, M.R.T. Lyu and R. Jin, "Discriminative semi-supervised feature selection via manifold regularization," \textit{IEEE Trans. Neural Netw.}, 21 (7) (2010) 1033-1047.
		
		\bibitem{Sun2021}N. Sun, T. Luo, W. Zhuge, H. Tao, C. Hou and D. Hu, "Semi-supervised learning with label proportion," \textit{IEEE Trans. Knowl. Data Eng.}, 35 (1) (2021) 877-890.
		
		\bibitem{semi-supervised2}X. Chang, F.Nie, Y. Yang and H. Huang, "A convex formulation for semi-supervised multi-label feature selection," \textit{Proc. AAAI Conf. Artif. Intell.}, 2014, pp. 1171-1177.
		
		\bibitem{LEDA2019}B. Guo, H. Tao, C. Hou and D. Yi, "Semi-supervised multi-label feature learning via label enlarged discriminant analysis," \textit{Knowl. Inf. Syst.}, 62 (2020) 2383-2417.
		
		\bibitem{SFAM2021}S. Lv, S.F. Shi, H.Z. Wang and F. Li, "Semi-supervised multi-label feature selection with adaptive structure learning and manifold learning," \textit{Knowl. Based Syst.}, 214 (2021) 106757.
		
		\bibitem{WLiu2021}W. Liu, H. Wang, X. Shen and I.W. Tsang, "The emerging trends of multi-label learning," \textit{IEEE Trans. Pattern Anal. Mach. Intell.}, 44 (11) (2021) 7955-7974.
		
		\bibitem{Laporte2013}L. Laporte, R. Flamary, S. Canu, S. Déjean and J. Mothe, "Nonconvex regularizations for feature selection in ranking with sparse SVM," \textit{IEEE Trans. Neural Netw. Learn. Syst.}, 25 (6) (2013) 1118-1130.
		
		\bibitem{YLiu2020}Y. Liu, D. Ye, W. Li, H. Wang and Y. Gao, "Robust neighborhood embedding for unsupervised feature selection," \textit{Knowl. Based Syst.} 193 (2020) 105462.
	
		\bibitem{SCFS2018}Y. Xu, J. Wang, S. An, J. Wei and J. Ruan, "Semi-supervised multi-label feature selection by preserving feature-label space consistency," \textit{Proc. 27th ACM Int. Conf. Inf. Knowl. Manag.}, 2018, pp. 783-792.
		
		\bibitem{SMLFS2021}Y. Zhang, Y. Ma, X. Yang, H. Zhu and T. Yang, "Semi-supervised multi-label feature selection with local logic information preserved," \textit{Adv. Comput. Intell.}, 1 (7) (2021) 1-15.
		
		\bibitem{SMDRIC2023}R. Li, J. Du, J. Ding, L. Jia, Y. Chen and Z. Shang, "Semi-supervised multi-label dimensionality reduction learning by instance and label correlations", \textit{Mathematics}, 11(3) (2023) 782.
		
		\bibitem{SFS-BLL2023}D. Shi, L. Zhu, J. Li, Z. Cheng and Z. Liu, "Binary label learning for semi-supervised feature selection," \textit{IEEE Trans. Knowl. Data Eng.}, 35 (3) (2023) 2299-2312.
		
		\bibitem{SMDR-MRC2024}R. Li, G. Zhou,  X. Li, L. Jia and Z. Shang, "Semi-supervised multi-label dimensionality reduction learning based on minimizing redundant correlation of specific and common features." \textit{Knowl. Based Syst.}, 294, 2024.
		
		\bibitem{SMFS2025}R. Sheikhpour, M. Mohammadi, K. Berahmand, F. Saberi-Movahed and H. Khosravi, "Robust semi-supervised multi-label feature selection based on shared subspace and manifold learning," \textit{Inf. Sci.}, 699 (2025) 121800.
		
		\bibitem{SFGR2025} D. Qing, Y. Zheng, W. Zhang, W.  Ren, X. Zeng and G. Li, "Semi-supervised feature selection with minimal redundancy based on group optimization strategy for multi-label data," \textit{Knowl. Inf. Syst.}, 67 (2) (2025) 1271-1308.
		
		
		\bibitem{SXiang2012}S. Xiang, F. Nie, G. Meng, C. Pan and C. Zhang, “Discriminative least squares regression for multiclass classification and feature selection”, \textit{IEEE Trans. Neural Netw. Learn. Syst.}, 23 (11) (2012) 1738-1754.
		
		\bibitem{XChen2017}X. Chen, F. Nie, G. Yuan and J.Z. Huang, “Semi-supervised feature selection via rescaled linear regression”, \textit{Proc. AAAI Int. Joint Conf. Artif. Intell.}, 2017, pp. 1525-1531.
		
		\bibitem{Lagrange}D.P. Bertsekas, "Constrained optimization and Lagrange multiplier methods," \textit{Academic Press}, 2014.
		
		\bibitem{FNie2010}F. Nie, H. Huang, X. Cai, and C. H. Ding, "Efficient and robust feature selection via joint l2,1-norms minimization", \textit{Proc. Adv. Neural Inf. Proc. Syst.}, 23 (2010) 1813-1821.
		
		\bibitem{JinHuang2014}J. Huang, F.P. Nie, H. Huang and C. Ding, "Robust manifold nonnegative matrix factorization," \textit{ACM Trans. Knowl. Disc. Data}, 8 (3) (2014) 1-21.
		
		\bibitem{F-solve1}R. Bartels and G. Stewart, "Algorithm 432 [C2]: Solution of the matrix equation AX+ XB= C [F4]," \textit{Commun. ACM}, 15 (9) (1972) 820-826.
		
		\bibitem{SMILE2017}Q. Tan, Y. Yu, G. Yu and J. Wang, "Semi-supervised multi-label classification using incomplete label information," \textit{Neurocomputing}, 260 (2017) 192-202.
		
		\bibitem{FSSRDM2018}L. Jiang, J. Wang and G. Yu, "Semi-supervised multi-label feature selection based on sparsity regularization and dependence maximization," \textit{Int. Conf. Intell. Control Inf. Proc.}, 2018, pp. 325-332.
		
		
		\bibitem{LSMR2020}V. Kraus, K. Benabdeslem and B. Canitia, "Laplacian-based semi-supervised multi-label regression," \textit{Int. Joint Conf. Neural Netw.}, 2020, pp. 1-8.
		
		\bibitem{UAFSBH2023}D. Shi, L. Zhu, J. Li, Z. Zhang and X. Chang, "Unsupervised adaptive feature selection with binary hashing," \textit{IEEE Trans. Image Process.}, 32 (2023) 838-853.
		
		\bibitem{SemiSetting}X. Chen, G. Yuan, F. Nie and Z. Ming, "Semi-supervised feature selection via sparse rescaled linear square regression," \textit{IEEE Trans. Knowl. Data Eng.}, 32 (1) (2020) 165-176.	
		
		\bibitem{MLKNN}M.L. Zhang and Z.H. Zhou, "ML-KNN: A lazy learning approach to multi-label learning," \textit{Pattern Recognit.}, 40 (7) (2007) 2038-2048.		
		
	\end{thebibliography}
\end{document}